%% file: main_draft.tex
\newif\ifconfver
\confverfalse      
\confvertrue        

\newif\ifplainver  
\plainvertrue
\plainverfalse

\newif\ifhide  
\hidetrue

\ifplainver
    \confverfalse   
\fi

\ifconfver
     \documentclass[10pt,twocolumn,twoside]{IEEEtran}
\else
    \ifplainver
        \documentclass[11pt]{article}
        \usepackage{fullpage}
    \else
        \documentclass[12pt,draftcls,onecolumn]{IEEEtran}
    \fi
\fi

\usepackage{calc,amsfonts,amssymb,amsmath,bm,bbm,url,color,theorem,graphicx,cite,enumerate}
\usepackage{psfrag,subfigure,float,array,booktabs,bbold}
\usepackage{algorithm}
\usepackage{algorithmicx}
\usepackage{algpseudocode}
\usepackage{multirow,multicol}
\usepackage{marvosym,comment}
\usepackage{calligra,nicefrac}
\usepackage[normalem]{ulem}

\usepackage[dvipsnames]{xcolor}
\usepackage{colortbl}

\usepackage{shortcuts_OPT}
\usepackage{todonotes,threeparttable}

\input{sym_marco.tex}

\definecolor{orange}{RGB}{255,107,0}
\definecolor{green}{RGB}{62, 218, 27}

\def\blue{\color{blue}}
\def\red{\color{red}}

\DeclareMathAlphabet\mathbfcal{OMS}{cmsy}{b}{n}

\theorembodyfont{\rmfamily}

\newtheorem{Remark}{Remark}

\begin{document}

\bibliographystyle{IEEEtran}

\newcommand{\papertitle}{
Hyperspectral Unmixing Under Endmember Variability: A Variational Inference Framework}

\newcommand{\paperabstract}{
...}


\ifplainver

     \title{\papertitle}

     \author{
     Yuening LI
      }

     \maketitle

\else
    \title{\papertitle}

    \ifconfver \else {\linespread{1.1}} \rm \fi

    \author{Yuening Li, Xiao Fu, Junbin Liu, and Wing-Kin Ma
\thanks{
Y. Li, J. Liu, and W.-K. Ma are with the Department of Electronic Engineering, The Chinese University of Hong Kong (e-mail: yuening@link.cuhk.edu.hk; liujunbin@link.cuhk.edu.hk; wkma@ieee.org).
X. Fu is with the School of Electrical Engineering and Computer Science at Oregon State University (e-mail: xiao.fu@oregonstate.edu).

The work of Y. Li, J. Liu and W.-K. Ma was supported by a General Research Fund (GRF) of Hong Kong Research Grant Council (RGC) under Project ID CUHK 14203721.

The work of X. Fu was supported in part by the National Science Foundation under project NSF ECCS-2024058.}
    }

    \maketitle

    \ifconfver \else
        \begin{center} \vspace*{-2\baselineskip}
        \end{center}
    \fi

    \ifconfver
    \else
    \fi

    \ifconfver \else \IEEEpeerreviewmaketitle \fi

 \fi

\ifconfver \else
    \ifplainver \else
        \newpage
\fi \fi

\allowdisplaybreaks
\setlength{\belowdisplayskip}{6pt} \setlength{\belowdisplayshortskip}{6pt}
\setlength{\abovedisplayskip}{6pt} \setlength{\abovedisplayshortskip}{2pt}

\begin{abstract}

This work proposes a variational inference (VI) framework for hyperspectral unmixing in the presence of endmember variability (HU-EV). An EV-accounted noisy linear mixture model (LMM) is considered, and the presence of outliers is also incorporated into the model. Following the marginalized maximum likelihood (MML) principle, a VI algorithmic structure is designed for probabilistic inference for HU-EV.
Specifically, a patch-wise static endmember assumption is employed to exploit spatial smoothness and to try to overcome the ill-posed nature of the HU-EV problem.
The design facilitates lightweight, continuous optimization-based updates under a variety of endmember priors. Some of the priors, such as the Beta prior, were previously used under computationally heavy, sampling-based probabilistic HU-EV methods.
The effectiveness of the proposed framework is demonstrated through synthetic, semi-real, and real-data experiments.

\end{abstract}

\begin{IEEEkeywords}\vspace{-0.0cm}
    Hyperspectral unmixing, endmember variability, outlier, Beta distribution, 
   variational inference approximation
\end{IEEEkeywords}

\section{Introduction}

In hyperspectral images (HSIs), multiple materials can be present simultaneously in the same pixels, due to the relatively coarse spatial resolution of HSI sensors \cite{keshava2002spectral}.
The pixels are thus considered as mixtures of the materials' spectral signatures (\ie endmembers) \cite{Ma2014HU,Jose12}. 
{\it Hyperspectral unmixing} (HU) aims to uncover the endmembers and estimate their proportions (\ie abundances) in the pixels. 
Numerous HU algorithms under different mixture models were proposed in the past two decades; see, \eg \cite{Winter1999,iordache2011sparse,fu2016semiblind,miao2007endmember,Dias2009,fu2016robust,qian2016matrix,ding2023fast,wu2021probabilistic,dobigeon2009joint,su2019daen,palsson2018hyperspectral}.

\begin{figure}[t]
    \centering
    \includegraphics[width=0.95\linewidth]{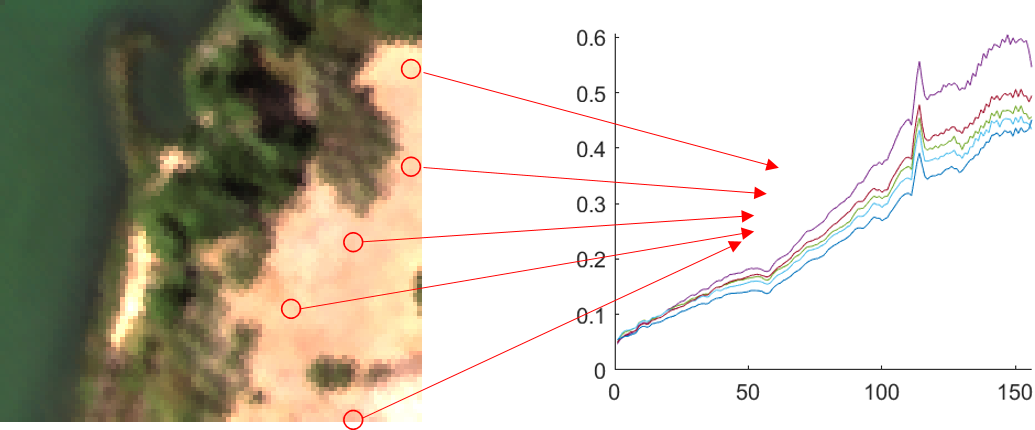}
    \caption{Illustration of EV in the Samson dataset. The spectra of the pixels that only contain ``{soil}'' are shown on the right.}
    \label{fig:EV}
\end{figure}

The existing mixture models and HU algorithms are effective to a large extent. Nonetheless, it was also noted that the widely used mixture models in HU, namely the {\it linear mixture model} (LMM) and its nonlinear extensions \cite{keshava2002spectral,Jose12,Ma2014HU},  may be over-simplified, especially in complex data acquisition environments \cite{Jose12}.
A main assumption in the classical LMM is that the endmembers are assumed to be unchanged across pixels.
However, the phenomenon called {\it endmember variability} (EV) has been observed in real-world HSIs \cite{roberts1998mapping,somers2011endmember,zare2013endmember}. That is, the endmembers' spectra vary from pixel to pixel, although they represent the same materials. EV arises due to many reasons, \eg the changes of atmospheric, illumination, and topographic conditions across pixels, as well as the intrinsic physicochemical differences of the material across the region of interest \cite{healey1999models,somers2011endmember,zare2013endmember,borsoi2021spectral}. An example of the EV effect is shown in Fig.~\ref{fig:EV}, where the spectra of the visually identified pixels representing ``{soil}'' are seen to vary across pixels. Such variations can degrade the performance of the HU algorithms that assume static endmembers.

Various approaches were proposed to tackle the {\it HU problem in the presence of EV} (HU-EV). For example, the works \cite{drumetz2016blind,thouvenin2015hyperspectral,drumetz2020spectral} used the concept of ``local endmembers'' in addition to a ``global'' reference to formulate a regularized model fitting (MF) criterion.
The works in \cite{bioucas2010alternating,drumetz2019hyperspectral,deng2013spatially,fu2016semiblind} constructed a dictionary of endmember variations, and formulated the HU-EV problem as a sparse regression (SR) problem. 
The performance of these methods is limited by a suite of factors, \eg the choice of the regularization functions, the regularization parameters, and/or the accuracy of the dictionary.
Another line of work takes a probabilistic perspective, treating the endmembers in each pixel as realizations of a certain distribution \cite{stein2003application,eches2010bayesian,halimi2015unsupervised,kazianka2011bayesian,woodbridge2019unmixing,zhou2018gaussian,li2020stochastic,du2014spatial}. 
By incorporating appropriate endmember/abundance priors,
HU-EV is then tackled by sophisticated statistical learning techniques, \eg Bayesian inference \cite{eches2010bayesian,halimi2015unsupervised,kazianka2011bayesian} and expectation-maximization (EM) \cite{stein2003application,woodbridge2019unmixing,zhou2018gaussian}.
Compared to the MF and SR-based approaches, the probabilistic approaches can lift us from the agony of parameter tuning by selecting proper priors, while maintaining reasonable performance, due to its adherence of the disciplined probabilistic inference techniques.

As a unique feature, the probabilistic approach embeds ``variability'' in the models. Such approach has attracted considerable attention in HU-EV; see, \eg \cite{eches2010bayesian,halimi2015unsupervised,kazianka2011bayesian,stein2003application,woodbridge2019unmixing,zhou2018gaussian}.
However, some challenges remain. Many probabilistic methods rely on algorithms that are either computationally demanding or prior-limited. A large portion of the methods take the Bayesian inference route, which often resorts to sampling operations to evaluate intractable integrals in the process (see, \eg \cite{eches2010bayesian,halimi2015unsupervised,kazianka2011bayesian}), hindering the scalability of the algorithms.
The EM-based methods  (see, \eg \cite{stein2003application,woodbridge2019unmixing,zhou2018gaussian}) are more efficient, but the viability of EM hinges on limited choices of endmember priors, \eg Gaussian or Gaussian mixtures.
 Gaussian-type distributions may not be the best fit for endmember models.
For example, the work \cite{du2014spatial} argued that the support of the endmembers is always nonnegative and bounded due to the pixel acquisition physics, and thus the Beta distributions are considered more suitable to serve as endmember priors.
 In the existing developments, there appears to have a dilemma between prior flexibility and computational efficiency.

Our interest lies in developing a probabilistic inference algorithmic framework for HU-EV, aiming to strike a balance between model flexibility and algorithm efficiency. 
To begin with, we propose an EV-accounted noisy LMM for hyperspectral pixels. The presence of outlying pixels is also taken into consideration, as pixels disobeying the LMM are widely observed \cite{dobigeon2013nonlinear,fu2016robust}.
Our idea is to formulate the endmember/abundance estimation problem as a {\it marginal maximum likelihood} (MML) criterion and design a {\it variational inference} (VI) algorithm to handle the MML objective. The VI approach recasts intractable estimation criteria (in terms of integral evaluation) as continuous optimization surrogates, allowing efficient updates \cite{wu2021probabilistic}. In addition, the VI's computing mechanism allows incorporating various priors, showing flexibility \cite{bishop2006pattern,blei2017variational}. However, directly applying VI under our HU-EV model leads to an under-determined estimation problem, hindering the validity of the whole process.
We propose to exploit spatial smoothness of the endmembers, \ie by assuming that the endmembers are patch-wise static, to circumvent this issue. 
To design a valid VI algorithm under this model, we derive a continuous optimization surrogate by sequentially applying Jensen's inequality under both patch and pixel scales.
The proposed VI framework can readily include various endmember priors (e.g., the Gaussian and Beta distributions) under its umbrella, all of which admit lightweight updates.

\smallskip

\noindent {\bf Notation.} 
The notations $x$, $\bx$ and $\X$ represent a scalar, a vector, and a matrix, respectively.
The symbols $\Rbb$ and $\Rbb_+$ denote the set of real numbers and non-negative numbers, respectively;
$\bx_i$ denotes the $i$th column of a matrix $\bX$;
$\Diag(\bx)$ denotes a diagonal matrix with diagonal elements given by $\bx$;
$\Diag(\bX)$ denotes a diagonal matrix collecting diagonal elements of $\bX$;
$\diag(\bX)$ denotes a vector containing diagonal elements of $\bX$;
$\| \cdot\|_2$ and $\|\cdot\|_F$ denote the Euclidean and Frobenius norm, respectively;
$\innerF \bA \bB = \sum_{i,j} A_{ij}B_{ij}$ denote the Frobenius inner product;
$\frac{\bx}{\by}$ and $\bx \odot \by$ denote the elementwise division and product, respectively;
$\bx \ge \by$ denotes the elementwise inequality;
$\bzero$ denotes an all-zero vector;
$\bone$ denotes an all-one vector;
$\bI$ denotes an identity matrix;
$\indfn{}$ denotes an indicator function;
$|\setI|$ denotes the number of elements in the set $\setI$;
$\Delta=\{ \bs\in\RR^N_+ | \bone^\top\bs = 1 \}$ denotes a unit simplex; 
$\setN(u,\sigma^2)=\frac{1}{\sqrt{2\pi \sigma^2}} \exp(-\frac{(x-u)^2}{2\sigma^2})$ denotes the PDF of $(u,\sigma^2)$-parameterized Gaussian distribution;
$\Beta(c,d)=\frac{1}{B(c,d)}x^{\alpha-1}(1-x)^{\beta-1}$ denotes the PDF of $(c,d)$-parameterized Beta distribution;
$ \Dir(\balp) = \frac{1}{B(\balp)} \left( \prod_{i=1}^{N} x_i^{\alpha_i - 1} \right) \indfn_{\Delta}(\bx)$ denotes the PDF of $\balp$-parameterized Dirichlet distribution;
$B(c,d) = \frac{\Gamma(c)\Gamma(d)} { \Gamma(c+d) }$ denotes the Beta function, and
$B(\balp) = \frac{\prod_{i=1}^N \Gamma(\alpha_i)} { \Gamma(\sum_{j=1}^N \alpha_j) } $ denotes the multivariate Beta function.

\section{Background and Preliminaries}

\subsection{HU and LMM}\label{sec:HU and LMM}
We begin with the {\it linear mixture model} (LMM), the most commonly used model in the literature of hyperspectral unmixing (HU) \cite{keshava2002spectral,Jose12,Ma2014HU}. In LMM, each pixel of a hyperspectral image is postulated to be a weighted sum of the endmembers, specifically,
\begin{align}\label{eq:simplex model}
\by_t = \bA\bs_t+\bv_t, ~~t=1,\dots,T.
\end{align}
Here, $\by_t \in \Rbb^M$ represents the $t$th pixel of the hyperspectral image, with $y_{i,t}$ being the measurement of a particular spectral band; $M$ is the number of spectral bands; $\bA =[\ba_1,\dots,\ba_N] \in \Rbb^{M\times N} $ is called the endmember matrix wherein $ \ba_i $ describes the spectral response of the $ i $th endmember; $N$ is the total number of endmembers; $ \bs_t \in \Delta$ is called the abundance vector of the $t$th pixel which describes the proportional distribution of the different endmembers at the $t$th pixel; $ \bv_t $ is noise.
The problem of HU is to recover $\bA$ and $\bs_1,\dots,\bs_T$ from $\by_1,\dots,\by_T$, which allows us to identify the underlying materials of the HSI and their corresponding proportions in each pixel. One should bear in mind that LMM is based on simplified assumptions. In particular, each endmember is assumed to be invariant across pixels.

\subsection{Preliminaries: Probabilistic HU via Variational Inference}
\label{sec:HU-VI}
There are various frameworks for solving the HU problem (see \eg \cite{Jose12}, for a survey). Among them are the probabilistic frameworks.
These frameworks work by modeling noise and the abundance vectors (and perhaps also the endmember matrix) as random quantities. Bayesian or maximum-likelihood (ML) inference techniques are then applied to estimate the endmember matrix and abundance vectors \cite{wu2021probabilistic,dobigeon2009joint}. Here we are interested in the marginal ML (MML) framework \cite{wu2021probabilistic}, where the noise and the abundance vectors are modeled as
\begin{align}\label{eq:asm}
    \bs_t \sim  {\rm Dir}(\bone) \quad\text{and}\quad
    \bv_t \sim \setN(\bzero,\sigma^2\bI),
\end{align}
respectively. Here, ${\rm Dir}(\balp)$ denotes the Dirichlet distribution with parameter $\balp = \bone$ and note that $\Dir(\bone)$ is the uniform distribution on the simplex $\Delta$. We consider the MML estimator
\begin{align}\label{eq:ML}
\btheta_{\rm ML} = \arg\max_{\btheta} \frac{1}{T}\sum_{t=1}^T \log  p_\btheta(\by_t),
\end{align}
where $\btheta=\{\bA,\sigma^2\}$ is the model parameter,
\begin{equation}
    \label{eq:marginal s}
    p_\btheta(\by_t) = \int p_\btheta(\by_t|\bs_t)p(\bs_t)  d\bs_t
\end{equation}
is the marginal likelihood, $p(\bs_t) \sim \Dir(\bone)$ is the prior distribution of $\bs_t$, and $p_\btheta(\by_t|\bs_t) \sim \setN(\bA\bs_t,\sigma^2\bI)$ is the likelihood of $\by_t$ given $\bs_t$. The terminology ``marginalized ML'' comes from the fact that we marginalize the likelihood with respect to $\bs_t$.

The MML criterion in \eqref{eq:ML} is difficult to deal with because the marginal likelihood \eqref{eq:marginal s} is in general an intractable integral. Importance (or Monte-Carlo) sampling schemes can be used to evaluate the integral in \eqref{eq:marginal s} in an approximate fashion,  which thereby allows us to apply numerical optimization to the MML problem \eqref{eq:ML} \cite{wu2021probabilistic}. It is worth noting that sampling was also used in other probabilistic frameworks for HU \cite{dobigeon2009joint}---once again for coping with some intractable integrals. However, it can be computationally demanding to use sampling when the size of the random variable to be sampled is not small \cite{wu2021probabilistic}.

{\it Variational inference} (VI) \cite{bishop2006pattern} is an approach that  approximates the marginal likelihood by a tractable expression, which thereby provides the opportunity to build efficient algorithms for handling the MML problem \eqref{eq:ML}. To describe the principle, let $q_t(\bs_t)$ be some given distribution with support $\Delta$. By Jensen's inequality,
\begin{equation}
\begin{aligned}\label{eq:jen}
\log p_\btheta(\by_t)
& = \log\left(  \Exp_{\bs_t \sim q_t}[ p_\btheta(\by_t|\bs_t)p(\bs_t)/q(\bs_t) ] \right)  \\
& \geq \Exp_{\bs_t\sim q_t}[ \log( p_\btheta(\by_t|\bs_t)p(\bs_t)/q(\bs_t) ) ]\\
&:= \hl(\btheta,q_t;\by_t),
\end{aligned}
\end{equation}
where $\Exp_{\bs_t\sim q_t}[\cdot]$ denotes expectation over $\bs_t$ with distribution $q_t(\bs_t)$.
The function $\hl$ in the above equation is called an {\it evidence lower bound} (ELBO) in the literature \cite{bishop2006pattern}, and the given distribution $q_t(\bs_t)$ is called a variational posterior. The idea of VI is to use the ELBO as a surrogate of the marginal likelihood, and, at the same time, to find a variational posterior that gives a tractable ELBO.

To understand how VI is used to tackle \eqref{eq:ML},
it would be easier to consider the specific form of $q_t$ used in the work \cite{wu2021probabilistic}. There, a Dirichlet variational posterior is used, \ie
\[
q_t(\bs_t)=\Dir(\balp_t).
\]
It can be shown that 
\begin{align*}\label{eq:l via}
& \hl(\btheta,q_t;\by_t) = \hl(\btheta,\bm \alpha_t;\by_t) \\
&\propto  -\frac{1}{2\sigma^2}
\big( \tr(\bA\bC_s(\balp_t)\bA\!^\top\!) -\! 2\by_t\!^\top\!\bA\bmu_s(\balp_t)\big) \!+\! H_s(\balp_t),
\end{align*}
where 
\begin{subequations}\label{eq:moments of s}
\begin{align}
    \bmu_s(\balp_t) &=\Exp_{\bs_t\sim q_t}[\bs_t]=\frac{\balp_t}{\bone^\top\balp_t}, \\
    \bC_s(\balp_t) &= \Exp_{\bs_t\sim q_t}[\bs_t\bs_t^\top] = \frac{\Diag(\balp_t) + \balp\balp^\top}{(1+\bone^\top\balp_t)\bone^\top\balp_t}, \\
    H_s(\balp_t) \!&=\! \log \!B(\balp_t) \!- \!(\balp_t\!-\!\bone)\!^\top\! (\psi(\balp_t)-\psi(\bone\!^\top\!\balp_t))
\end{align}
\end{subequations}
are the first, second moments and entropy of $\bs_t$, respectively, and $\psi$ is the digamma function. Consequently  one can approximate the MML problem \eqref{eq:ML} by 
\begin{align}\label{eq:approx ML prob}
\max_{\btheta,\bphi}
\hL(\btheta, \bphi;\bY):= 
\frac{1}{T} \sum_{t=1}^T \hl(\btheta,\balp_t;\by_t),
\end{align}
where $\bphi=\{\balp_1,\dots,\balp_T\}$. In particular, the variational posterior parameters $\bphi$ are optimized to give the best ELBO under Dirichlet variational posteriors.

\begin{figure}[t]
\centering
\includegraphics[width=0.7\linewidth]{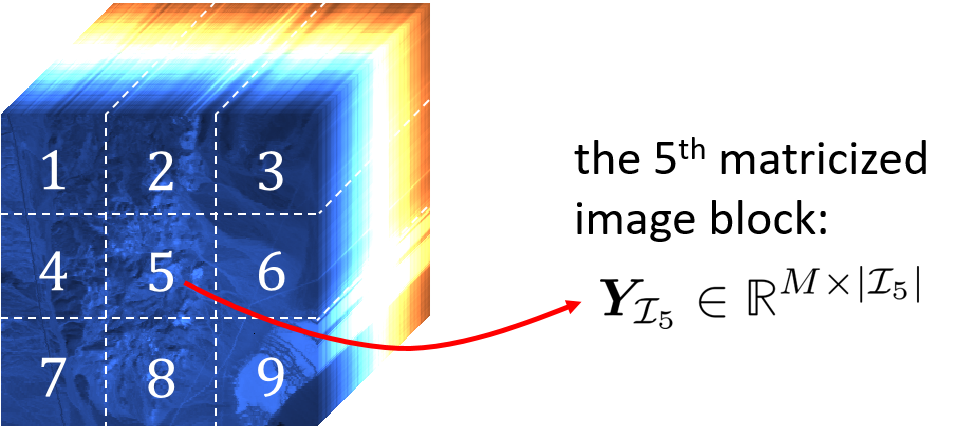}
\caption{Segmentation of an image into blocks. $|\setI_5|$ denotes the number of pixels of 5-th block.}
\label{fig:segmentation}
\end{figure}

\begin{figure}[t]
\centering
\subfigure[Samson image]{
    \label{fig:samson}
    \includegraphics[width=0.33\linewidth]{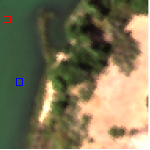} }
\quad
\subfigure[Spectra in two areas]{
    \label{fig:samson spectra}
    \includegraphics[width=0.45\linewidth]{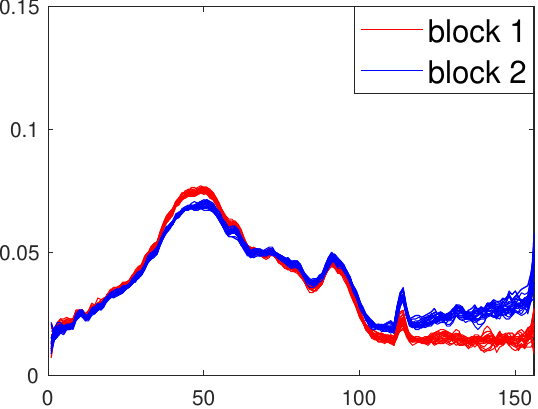} }
\subfigure[Moffett image]{
    \label{fig:moffett}
    \includegraphics[width=0.33\linewidth]{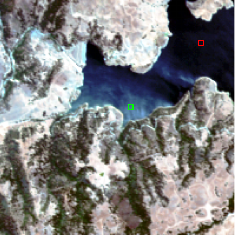} }
\quad
\subfigure[Spectra in two areas]{
    \label{fig:moffett spectra}
    \includegraphics[width=0.45\linewidth]{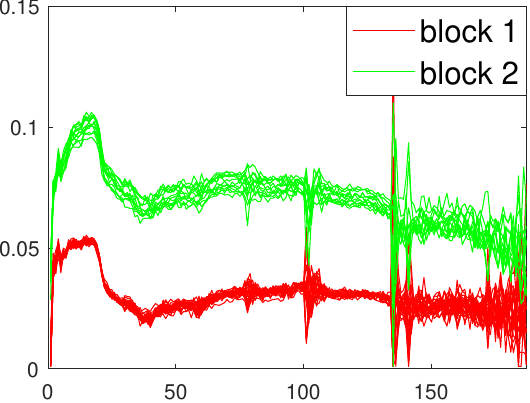} }
\caption{Illustration of spatial smoothness of the endmembers in the presence of EV. 
The two manually cropped blocks (with size $5\times 5)$ for the water in the Samson and Moffett images are shown in the left column. The spectra are shown in the right column. }
\label{fig:illustration EV}
\end{figure}

\section{Proposed Approach}

\subsection{Proposed HU-EV Model}
As described in Section \ref{sec:HU and LMM}, the LMM is built on the premise that each endmember does not vary across pixels. In practice, there are instances for which endmembers vary from pixel to pixel in intensity or in shape \cite{roberts1998mapping,somers2011endmember,zare2013endmember}. For such instances it is appropriate to modify the LMM in \eqref{eq:simplex model} as 
\begin{equation}
    \by_t = \bA_t\bs_t + \bv_t,
\end{equation}
where $\bA_t$ is the endmember matrix at the $t$th pixel. The challenging aspect lies in how we model $\bA_t$. Many models in the literature (\eg \cite{drumetz2016blind,thouvenin2015hyperspectral,drumetz2020spectral}) may be regarded as taking the form 
\begin{equation}\label{eq:At form}
    \bA_t = \bA\bC_t + \bE_t,
\end{equation}
where $\bC_t=\Diag(c_{1,t},\dots,c_{N,t})$ is a diagonal matrix that alters the scalings of $\ba_1,\dots,\ba_N$, $\bE_t$ is a perturbation term, and $\bA$ represents the reference endmember matrix. For example, the extended LMM has $\bE_t=\bzero$ \cite{drumetz2016blind}; the perturbed LMM has $\bC_t=\bI$ \cite{thouvenin2015hyperspectral}; the model in \cite{drumetz2020spectral} considers the more general form in \eqref{eq:At form}. In the existing literature, it is common to embed spatial smoothness of the endmember variations in the model \cite{drumetz2016blind,thouvenin2015hyperspectral}. For instance, one may promote spatial smoothness among each $\{ c_{i,t} \}_{t=1}^T$ across all $t$'s via regularization \cite{drumetz2016blind}.

Our model is as follows. We model $\bA_t$ as being block-wise invariant, namely, \vspace{-1em}
\begin{align}\label{eq:At Ak}
    \bA_t = \bA_k, ~~\forall t\in \setI_k,
\end{align}
where $\setI_k$ denotes the pixel index set of the $k$th patch;  see Fig. \ref{fig:segmentation} for an illustration. This means that we assume that the endmembers do not change within a local region, which may be seen as a way to exploit spatial smoothness. 
Fig.~\ref{fig:illustration EV} shows illustrations using the Samson and Moffett data, respectively. In Samson, the spectra of water in the red and blue blocks are seen to have differences---yet the water spectra within the blocks are fairly similar. This phenomenon is also observed in Moffett.
We model each patch-wise invariant endmember $\bA_k$ as a random quantity; \ie 
\begin{align}\label{eq:dist_Ak}
    \bA_k\sim p_{\btheta_A}(\bA_k),
\end{align}
where $p_{\btheta_A}(\bA_k)$ is the prior of $\bA_k$ with model parameter $\btheta_A$. We can model $\bA_k$ as Gaussian \cite{stein2003application,eches2010bayesian,li2020stochastic,halimi2015unsupervised,kazianka2011bayesian,woodbridge2019unmixing}. Or, we can model $\bA_k$ as Beta distributed \cite{du2014spatial}; the support of endmembers is $[0,1]$ and hence the Beta distribution model seems better in modeling the $[0,1]$ support nature---more options for $ p_{\btheta_A}(\bA_k)$ that can be handled in our framework are summarized in Table~\ref{tab:support dist}. 
Also, we assume $\bm s_t\sim {\rm Dir}(\bm 1)$ and ${\bm v}_t\sim {\cal N}(\bm 0,\sigma^2 \bm I)$.

We also take outliers into consideration. Specifically, our signal model for $\by_t$ is given by
\begin{align}\label{eq:ourmodel}
    \by_t = \begin{cases}
         \bA_k \bs_t + \bv_t,\quad &{\rm w.p.}~1-\gamma,\\
         \bo_t,\quad &{\rm w.p.}~\gamma,
    \end{cases}~\forall t\in {\cal I}_k.
\end{align}
The above means that we assume the probability of each pixel being an outlier $\bo_t$ to be $\gamma$. The outliers refer to the pixels that do not obey the ``nominal model'' $\by_t=\bA_k \bs_t + \bv_t$ (\eg
those caused by nonlinear mixing effects \cite{dobigeon2013nonlinear} and/or heavily corrupted measurements \cite{syu2019outlier}).
To reflect our uncertainties with outliers, we model the outlier as a large random quantity  
\begin{align}\label{eq:outlier_dist}
\bm o_t\sim p_{\rm out}(\by_t),    
\end{align}
where $p_{\rm out}(\by_t)$ is a pre-specified distribution.
We should note that using such a mixture distribution to model nominal samples/outliers has a long history in machine learning and signal processing; see \cite{eskin2000anomaly,aitkin1980mixture,lauer2001mixture} and applications in HU \cite{li2021robust,wu2017stochastic}.

\vspace{-1em}

\subsection{MML and VI for HU-EV}
In this subsection, we propose to handle the HU-EV problem under the model in \eqref{eq:ourmodel} using VI.
To be specific, consider the $k$th patch. We have
\begin{equation}\label{eq:marginal Yk outlier}
     p_\btheta(\bY_{\setI_k}) = \int \prod_{t\in\setI_k}  p_{\btheta}(\by_t | \bA_k) p_{\btheta_A}(\bA_k) d\bA_k,
\end{equation}
in which 
\begin{align}\label{eq:marginal yt outlier}
p_\btheta(\by_t|\bA_k) &= \gamma \cdot \pout(\by_t) \\
&+ (1-\gamma)\cdot
\underbrace{\int p_{\btheta}(\by_t | \bA_k,\bs_t) p(\bs_t) d\bs_t}_{\pnom(\by_t|\bA_k)}, ~\forall t\in {\cal I}_k,\nonumber
\end{align}
where the last term is the conditional distribution of $\bm y_t$ given $\bA_k$ and given that $\by_t$ is a nominal data. 
Note that we define $\btheta= \{\btheta_A,\sigma^2,\gamma\}$.
The MML criterion is given by
\begin{align}\label{eq:ML bcm}
    \btheta_{\rm ML} = \arg \max_{\btheta} \setL(\btheta;\bY):=
  \sum_{k=1}^K  \log p_{\btheta}(\bY_{\setI_k}).
\end{align}
Again, the likelihood $p_{\btheta}(\bY_{\setI_k})$ is hard to express analytically. Notice that the way of constructing ELBO in the previous section no longer applies---the fact that both $\bA_k$ and $\bs_t$ are random variables with different ``spatial scales'' (\ie patch v.s. pixel) does not allow us to use a single Jensen's inequality to find a manageable ELBO.
The existence of the outlier component in \eqref{eq:ourmodel} also needs to be taken into consideration.

To move forward, we propose to employ the mean-field trick in VI \cite{bishop2006pattern}. The idea is to treat the variational posteriors of $\bA_k$ and $\bs_t$ separately. To be specific, assume that we have picked $q_k(\bA_k)$ as the variational posterior of $\bA_k$. Then we have
\begin{equation}
\begin{aligned}\label{eq:jen A}
	\log p_\btheta(\bY_{\setI_k})
	&= \log \Exp \Big[ \prod_{t\in\setI_k} p_{\btheta}(\by_t | \bA_k) p_{\btheta_A}(\bA_k) / q_k(\bA_k) \Big]  \\
	& \geq  \Exp \Big[ \sum_{t\in\setI_k} \log  p_{\btheta}(\by_t | \bA_k) + \log \frac{p_{\btheta_A}(\bA_k) }{q_k(\bA_k)}  \Big], 
\end{aligned}
\end{equation}
where we used $\Exp[\cdot]$ as a shorthand notation for $\Exp_{\bA_k \sim q_k}[\cdot]$. 
We further bound $\log p_{\btheta}(\by_t|\bA_k )$ by 
\begin{align}\label{eq:jen out}
    &\log  p_\btheta  (\by_t|\bA_k\!) 
    = 
    \log\! \Big( \! \gamma \!\cdot \!\pout(\by_t) \!+ \! (1\!-\!\gamma) \pnom(\by_t|\bA_k ) \! \Big) \\
     &~~~\ge \omega_t \log \! \frac{\gamma \!\cdot\! \pout(\by_t)}{\omega_t} + \bomg_t  \Big(\! \log \!\frac{1\!-\!\gamma}{\bomg_t} \!+ \!\log \pnom(\by_t|\bA_k )    \!\Big), \nonumber
\end{align}
where $0\le \omega_t\le 1, \bomg_t=1-\omega_t$.
The $\omega_t$'s are parameters to learn in our VI framework.
The term $ \log \pnom(\by_t|\bA_k ) $ can be lower bounded by the Jensen's inequality
 \begin{align}\label{eq:jen nom}
    \!\! \log \pnom(\by_t|\bA_k ) \!
    &= \log \Exp_{\bs_t\sim q_t}[ p_{\btheta}(\by_t|\bA_k,\bs_t)p(\bs_t)/q_t(\bs_t)  ] \nonumber\\
    \ge \Exp_{\bs_t\sim q_t} &\Big[\log p_\btheta(\by_t|\bA_k,\bs_t) + \log \frac{p(\bs_t)}{q_t(\bs_t)}  \Big] 
\end{align} 
where $q_t(\bs_t)$ is the variational posterior of $\bs_t$, which is chosen to follow the ${\rm Dir}(\bm \alpha_t)$ distribution as in \cite{wu2021probabilistic} (see Sec. \ref{sec:HU-VI}).

The lower bound in \eqref{eq:jen nom}, after taking expectation over $\bA_k$, can be expressed more explicitly as follows:
\begin{align}\label{eq:hltk}
& \hl_{t,k}(\sigma^2\!,\bphi_k^A,\balp_t;\by_t)\\
    & = \!
   \frac{1}{2\sigma^2} \!\Big( \! 2\by_t^\top \bmu_A(\bphi_k^A) \bmu_s(\balp_t) \!- \tr\big({\bC}_A(\bphi_k^A){\bC}_s(\balp_t) \big) \!-\!\by_t^\top\!\by_t  \!\Big)  \nonumber\\
    &\qquad - \! \frac{M}{2}\!\log\sigma^2 +  H_s(\balp_t)   + {\rm const} , \nonumber
\end{align}
where $\bphi_k^A$ is the parameter of the variational posterior $q_k(\bA_k)$, 
$\bmu_A(\bphi_k^A) = \Exp[\bA_k] $ and
$\bC_A(\bphi_k^A) = \Exp[\bA_k^\top\bA_k] $ are the first- and second-order moments of $\bA_k$, respectively ($\bmu_s(\balp_t)$ and $\bC_s(\balp_t)$ are defined similarly), and $H_s(\balp_t)$ is the entropy of $\bs_t$ as given in \eqref{eq:moments of s}. Using this notation,
one can find an explicit lower bound of the RHS of \eqref{eq:jen A}, i.e.,
\begin{align}\label{eq:logpthetaYk}
&\log p_\btheta(\bY_{\setI_k}) \\
&\geq  \underbrace{\!\!\sum_{t\in {\cal I}_k} \!\!\Big( \bomg_t \hl_{t,k}(\sigma^2\!,\bphi_k^A,\balp_t;\by_t) + g_t(\omega_t,\!\gamma) \!\Big) -  {\rm KL}_A(\btheta_A,\bphi_k^A)}
_{:=\hl_k(\btheta,\bphi;\bY_{\setI_k})},  \nonumber 
\end{align}
where we define $\bphi=\{ \bphi_k^A,\balp_t, \omega_t\}$, $\kl_A(\btheta_A,\bphi_k^A)=\Exp_{\bA_k\sim q_k} [\log \frac{q_k(\bA_k)}{p_{\btheta_A}(\bA_k)}]$ is the KL divergence between $q_k(\bA_k)$ and $p_{\btheta_A}(\bA_k)$, and 
\[ g_t(\omega_t,\gamma)=\omega_t \log  \frac{\gamma \cdot\pout(\by_t)}{\omega_t} + \bomg_t \log \frac{1-\gamma}{\bomg_t}. \nonumber \]
The above leads to an ELBO of $\setL(\btheta;\bY)$, \ie 
\[ \setL(\btheta;\bY)\ge\hL(\btheta,\bphi;\bY)=\sum_{k=1}^K 
\hl_k(\btheta,\bphi;\bY_{\setI_k}). \]
Like before, we find the parameters of interest via
\begin{align}\label{eq:ourobj}
    (\widehat{\btheta},\widehat{\bphi}) = \arg\max_{\btheta,\bphi} \hL(\btheta,\bphi;\bY).
\end{align}
We refer to this computational scheme as the {\it \underline{H}yperspectral \underline{E}ndmember \underline{L}earning under \underline{E}ndmember Variatio\underline{N}s} (\texttt{HELEN}) algorithmic framework.

\begin{Remark}
In our derivation, we used an abstract representation for $q_k(\bA_k)$.
In practice, $q_k(\bA_k)$ should be selected according two principles:
First, it should be chosen according to $p_{\btheta_A}(\bA_k)$, as $q_k(\bA_k)$ is meant to approximate the ground-truth posterior $p(\bA_k|{\bm Y}_{{\cal I}_k})$ and share the same support of $p_{\btheta_A}(\bA_k)$.
Second, $q_k(\bA_k)$ is expected to have analytical expressions for the mean $\bmu(\bphi_k^A)$, the correlation $\bm C(\bphi_k^A)$, and the KL-divergence $\kl_A(\btheta_A,\bphi_k^A)$ in \eqref{eq:logpthetaYk}. This way, \eqref{eq:ourobj} is relatively easy to tackle, preferably using readily available continuous optimization toolkit.
In the context of HU-EV, a variety of $p_{\btheta_A}(\bA_k)$ and $q_k(\bA_k)$ can serve these purposes. Table \ref{tab:support dist} shows a collection of them.
In the next two subsections, we will showcase detailed VI algorithm design under the VI principle using a couple of reasonable priors of $p_{\btheta_A}(\bA_k)$.
\end{Remark}

\begin{Remark}
The derivation also shows the importance of incorporating the patch-wise static model for $\bA_k$ [cf. Eq.~\eqref{eq:At Ak}]. As mentioned, the model is a reasonable reflection of the spatial smoothness of the endmembers in reality (see Fig.~\ref{fig:illustration EV}). More importantly, it helps us mitigate an intrinsic under-determined issue in the VI framework. To be specific, the VI framework has to estimate $\bm \phi^A_k$ for each $\bA_k$ in the process. If $|{\cal I}_k|$ is overly small (say, if $|{\cal I}_k|=d$ and $ {\rm size}(\bphi_k^A) > Md\,$, where $ {\rm size} (\bphi_k^A)$ denotes the number of elements of $\bphi_k^A$),
then we have to use $Md$ data measurements (\ie ${\bY}_{{\cal I}_k}$) to underpin $\bphi_k^A$, which is not sufficient in general.
This subtle point in fact stands as a fundamental hurdle for designing an effective VI algorithm to handle the HU-EV problem under \eqref{eq:ourmodel}---using \eqref{eq:At Ak} and a sufficiently large $|{\cal I}_k|$ should resolve the issue. On the other hand, $|{\cal I}_k|$ should not be overly large either, in order to capture sufficient variations of $\bA_k$ across patches and to accumulate a large enough sample size (\ie $K$) for meaningful inference of $\btheta_A$.
 Using the spatial smoothness assumption in \eqref{eq:At Ak} has another side benefit: Upon convergence, the VI algorithm will automatically output the estimations for $\mathbb{E}[\bA_k]=\bmu_A(\bphi^A_k)$, which can readily serve as the estimation for the endmembers in patch $k$. The collection $\{\bmu_A(\bphi^A_k) \}_{k=1}^K$ reveals the spatial profiles of the endmembers, which is important information in the context of hyperspectral analysis---yet many existing HU-EV methods lack the ability to estimate such profiles (see, e.g.,\cite{eches2010bayesian,halimi2015unsupervised,kazianka2011bayesian,li2020stochastic,stein2003application,woodbridge2019unmixing}).
\end{Remark}

\subsection{The Beta Prior Case}
In this subsection, we choose the Beta prior for $\bA_k$, and use this prior to showcase detailed algorithm design under the proposed \texttt{HELEN} framework. To be specific, we assume
\begin{align}\label{eq:BetaA}
\bA_k\sim  {\rm Beta}(\bm C,\bm D).
\end{align}
The above means $[\bA_k]_{ij}\sim {\rm Beta}(C_{ij},D_{ij})$ where $\bm C\in \mathbb{R}_{++}^{M\times N}$ and $\bm D\in \mathbb{R}_{++}^{M\times N}$; that is, we have $\btheta_A =\{\bC,\bD\}$.  The Beta prior is considered appropriate for modeling endmembers, as it has a nonnegative and bounded support \cite{du2014spatial}.

Following Table~\ref{tab:support dist}, we use $q_k(\bA_k)= \Beta(\bU_k,\bV_k )$. Under our notation in \texttt{HELEN}, $\bphi^k_A=(\bU_k,\bV_k )$.
Additionally, we have:
\begin{align*}
    \bmu_A(\bphi_k^A) &= \frac{\bU_k}{ \bU_k+\bV_k},\\
    \bC_A(\bphi^A_k) &= \bmu_A(\bphi_k^A)^\top \bmu_A(\bphi_k^A) + \Diag(\bmu_A(\bphi_k^A)^\top\tbV_k),\\
    {\rm KL}_A(\btheta_A,\bphi_k^A) 
    &=  \log \!B(\bC,\bD)  - \innerF {\bC\!-\!\bU_k}{\psi(\bU_k)} \nonumber\\
    &\quad~ - \!\log \!B(\bU_k,\bV_k) - \innerF{\bD\!-\!\bV_k}{\psi(\bV_k) } \\
    &\quad~ + \innerF{\bC\!+\!\bD\!-\!\bU_k\!-\!\bV_k} {\psi(\bU_k+\bV_k)},\nonumber
\end{align*}
where 
$\tbV_k = \frac{\bV_k}{ (\bU_k+\bV_k)\odot(\bU_k+\bV_k+1)}$, and $\odot$ denotes the Hadamard product.
By substituting the above into $\widehat{\ell}_{t,k}$ [cf. Eq.~\eqref{eq:hltk}],
the $\hL(\btheta,\bphi;\bY)$ in \eqref{eq:ourobj} is a continuous function.

To tackle the maximization problem of $\hL(\btheta,\bphi;\bY)$, many off-the-shelf optimization tools can be used. In this work, we propose a simple alternating maximization (AM) scheme as follows:
\begin{subequations}\label{eq:am beta}
 \begin{align}
    \balp_t  &\in \arg\max_{\balp_t\in\Rbb_{++}^N} \hl_{t,k}(\sigma^2\!,\bphi_k^A,\balp_t;\by_t), \forall t,\label{eq:prob alp}\\
     \bU_k,\bV_k &\in \arg\max_{\bU_k,\bV_k\in \Rbb_{++}^{M\times N}} 
     \hl_k(\btheta,\bphi;\bY_{\setI_k}),
     \forall k,\label{eq:prob UV}\\
     \omega_t &\in \arg \!\!\!\max_{0\le \omega_t\le 1} \bomg_t \hl_{t,k}(\sigma^2\!,\bphi_k^A,\balp_t;\by_t) + g_t(\omega_t,\!\gamma), \forall t,\label{eq:prob omg}\\
     \bC,\bD &\in \arg\max_{\bC,\bD\in \Rbb_{++}^{M\times N}} \textstyle
     -\sum_{k=1}^K {\rm KL}_A(\btheta_A,\bphi_k^A), \label{eq:prob CD}\\
    \sigma^2 &\in \arg\max_{\sigma^2}\textstyle
    \sum_{k=1}^K \sum_{t\in\setI_k} \bomg_t \hl_{t,k}(\sigma^2,\bphi_k^A,\balp_t;\by_t), \label{eq:prob sig}\\
    \gamma &\in  \arg\max_{\gamma} \textstyle \sum_{k=1}^K \sum_{t\in\setI_k} g_t(\omega_t,\!\gamma).\label{eq:prob gam}
\end{align}   
\end{subequations}
The expressions are obtained via discarding the terms that are constants after fixing the pertinent variables.

To see how the subproblems in \eqref{eq:am beta} are handled, first notice that the subproblems in \eqref{eq:prob omg}, \eqref{eq:prob sig} and \eqref{eq:prob gam} have closed-form solutions:
\begin{align}
    \omega_t &= \frac{\gamma \cdot \pout(\by_t)}{\gamma \cdot \!\pout(\by_t) + (1\! - \!\gamma)\cdot \exp(\hl_{t,k}(\sigma^2 \! , \bphi_k^A,\balp_t;\by_t) )},\label{eq:sol omg}\\
    \sigma^2 &= \frac{\sum_{k=1}^K  \epsilon_{k} }{M\times (\sum_{t=1}^T\bomg_t)}, \label{eq:sol sig outlier}\\
    \gamma &= \textstyle \frac{1}{T}\sum_{i=1}^T \omega_t,\label{eq:sol gam}
\end{align}
where $\epsilon_{k} = \sum_{t\in\setI_k} \bomg_t \Big( 2\by_t^\top \bmu_A(\bphi_k^A) \bmu_s(\balp_t) \!- \tr({\bC}_A(\bphi_k^A){\bC}_s(\balp_t) ) \!-\!\by_t^\top\!\by_t \Big)$.
The subproblems in \eqref{eq:prob alp}, \eqref{eq:prob UV} and \eqref{eq:prob CD} have nonconvex optimization objectives. 
Nonetheless, as they are differentiable, we propose to use {\it accelerated projected gradient} (APG) \cite{beck2017first} to update the parameters $\{\balp_t\}_{t=1}^T,\{\bU_k,\bV_k\}_{k=1}^K,\bC$ and $\bD$, where the step size is determined by the backtracking line search \cite{beck2017first}. The updates are summarized in Algorithm \ref{alg:beta}, which is referred to as \texttt{HELEN (Beta)}.
Notably, in principle, the Beta prior can be used in other probabilistic HU-EV framework, \eg those in \cite{eches2010bayesian,halimi2015unsupervised,kazianka2011bayesian}.
However, \texttt{HELEN (Beta)} is the first to offer an efficient, sampling-free solution, to our best knowledge.

\begin{algorithm}[t]
\renewcommand{\algorithmicrequire}{\textbf{Input:}}
\renewcommand{\algorithmicensure}{\textbf{Output:}}
\caption{\texttt{HELEN (Beta)}}
\label{alg:beta}
\begin{algorithmic}[1]
    \Require $\{\by_t\}_{t=1}^T$;
    \State {\bf initialize:} $ \bC,\bD,\sigma^2,\gamma , \{ \balp_t,\omega_t \}_{t=1}^T,\{\bU_k,\bV_k\}_{k=1}^K$;
    \Repeat	
    \State \textit{\% update variational parameters}
    \State update $ \balp_t$ by APG; 
    \State update $ \bU_k,\bV_k $ by APG;
    \myState {$ \omega_t=\frac{\gamma \cdot \pout(\by_t)}{\gamma \cdot \pout(\by_t) + (1-\gamma)\cdot \exp(\hl_{t,k}(\sigma^2,\bphi;\by_t) )} $ };   \State \textit{\% update model parameters}
    \State update $\bC,\bD$ by APG;
    \myState{$\sigma^2 = \frac{\sum_{k=1}^K  \epsilon_{k} }{M\times (\sum_{t=1}^T\bomg_t)}$};
    \myState  {$\gamma = \frac{1}{T}\sum_{i=1}^T \omega_t$ };
    \Until {a stopping rule is satisfied}
    \State \textit{\% calculate posterior mean of $\bA_k$ and $\bs_t$}
    \State $\hbA_k = \frac{\bU_k}{ \bU_k+\bV_k}, \hbs_t = \frac{\balp_t}{\bone^\top\balp_t} $; 
    \vspace{0.3em}
    \Ensure $\{ \hbA_k\}_{k=1}^K, \{\hbs_t,\omega_t\}_{t=1}^T$.
\end{algorithmic}
\end{algorithm}

\subsection{The Gaussian Prior Case}
To demonstrate the flexibility of the \texttt{HELEN} framework, we also derive an algorithm under the Gaussian prior, \ie
\begin{equation}\label{eq:gaussA}
    \bA_k \sim \setN(\bbA,\bQ),
\end{equation}
where $[\bA_k]_{ij} \sim \setN(\bar{A}_{ij},Q_{ij})$.
Here, $\btheta_A=\{ \bbA,\bQ \}$.
Compared to the Beta prior, the Gaussian prior is slightly less intuitive in the context of HU. This is because the Gaussian distribution has an unbounded support that could produce negative realizations---but hyperspectral measurements are all nonnegative.
Nonetheless, the Gaussian distribution is still considered useful in representing a large variety of data, by properly adjusting its mean and variance. In addition, as we will see shortly, using the Gaussian prior under the \texttt{HELEN} framework helps come up with more efficient updates.

By Table \ref{tab:support dist}, the variational posterior $q_k(\bA_k)$ is chosen as
\begin{align}
    q_k(\bA_k) = \setN(\bU_k,\bSig_k),
\end{align}
with $\bU_k\in\Rbb_{++}^{M\times N}$,
which means that $\bphi_{k}^A=\{\bU_k,\bSig_k\}$.
In this case, one can see that
\begin{align*}
    \bmu_A(\bphi_k^A) &= \bU_k,\\
    \bC_A(\bphi^A_k) &= \bU_k^\top\bU_k +\Diag(\bone^\top\bSig_k),\\
    \!\!{\rm KL}\!_A(\btheta\!_A,\!\bphi_k^A\!) \!\!
    &= \!\frac{1}{2}\!\!\sum_{m=\!1}^M \!\!\sum_{n=\!1\!\!}^N \!\!\frac{([\bU\!_k]_{mn} \!\!-\!\! \bbA_{mn}\!)\!^2 \!+\! [\bSig_k]_{mn} }{ \bQ_{mn} } \!\!+\! \log \!\frac{\bQ_{mn}}{[\bSig_k]_{mn}}\!.
\end{align*}
With the above expressions, we tackle the AM scheme in \eqref{eq:am beta}, with subproblem \eqref{eq:prob UV} and \eqref{eq:prob CD} replaced by
\begin{subequations}\label{eq:am gau}
\begin{align}
     \bU_k &\in \arg\max_{\bU_k\in \Rbb_{++}^{M\times N}} 
     \hl_k(\btheta,\bphi;\bY_{\setI_k}),
     \forall k,\label{eq:prob Uk gau}\\   \bSig_k &\in \arg\max_{\bSig_k\in \Rbb_{++}^{M\times N}} 
     \hl_k(\btheta,\bphi;\bY_{\setI_k}),
     \forall k,\label{eq:prob Sigk gau}\\
     \bbA,\bQ &\in \arg\max_{\bbA,\bQ\in \Rbb_{++}^{M\times N}} \textstyle
     -\sum_{k=1}^K {\rm KL}_A(\btheta_A,\bphi_k^A). \label{eq:prob pA gau}
\end{align}   
\end{subequations}
In the above, the updates of $\bm \Sigma_k$, $\bar{\bm A}$ and $\bm Q$ admit analytical forms:
\begin{subequations}\label{eq:sol gau}
\begin{align}
    \bSig_k &= \frac{1}{1/\bQ + \bone \diag(\bR_{\bs}^k)/\sigma^2}, \label{eq:sol qA Sig gau}\\
    \bbA &= \textstyle \frac{1}{K}\sum_{k=1}^K \bU_k,\\
    \bQ &= \textstyle \frac{1}{K}\sum_{k=1}^K (\bU_k-\bbA)^2+\bSig_k.\label{eq:sol pA gau}
    \end{align}
\end{subequations}
To update $\bm U_k$, we still use APG as in the Beta case.
However, the nice form of \eqref{eq:prob Uk gau} 
allows us to update $\bU_k$ by APG without using line search. To be specific, the objective function in \eqref{eq:prob Uk gau} takes the form:
\begin{align}\label{eq:obj l_Uk}
    \! \! \!\! \hl_k\! \propto \!\!\frac{1}{2\sigma^2} 
    \tr ( \bU_k^\top\!\bU_k \bR_s \!-\!  2\bY_s\bU_k \!) \! + \!\!\sum_{m=\!1}^M \!\!\sum_{n=\!1\!\!}^N \!\!\frac{([\bU\!_k]_{mn} \!\!-\!\! \bbA_{mn}\!)\!^2  }{ 2\bQ_{mn} },\! 
\end{align}
where $\bR_s = \sum_{t\in\setI_k}\bomg_t \bC_s(\balp_t)$, $\bY_s = \sum_{t\in\setI_k}\bomg_t\bmu_s(\balp_t)\by_t^\top$.
The gradient of \eqref{eq:obj l_Uk} with respect to $\bU_k$ is Lipschitz continuous and easy to acquire (i.e., $ L=\|\bR_s\|_2/\sigma^2 +  \|1/\bQ\|_F   $).
Hence, setting the step size as $1/L$ ensures that APG converges to the optimum of the subproblem \cite{beck2017first}.

We call this algorithm \texttt{HELEN (Gauss)} (see Algorithm~\ref{alg:gauss}). 
In summary, in \texttt{HELEN (Gauss)}, only the update of $\balp_t$ and $\bU_k$ need to call the iterative APG algorithm---leading to simpler updates relative to \texttt{HELEN (Beta)}.

\begin{algorithm}[t]
\renewcommand{\algorithmicrequire}{\textbf{Input:}}
\renewcommand{\algorithmicensure}{\textbf{Output:}}
\caption{\texttt{HELEN (Gauss)} }
\label{alg:gauss}
\begin{algorithmic}[1]
    \Require $\{\by_t\}_{t=1}^T$;
    \State {\bf initialize:} $ \bbA,\bQ,\sigma^2,\gamma , \{ \balp_t,\omega_t \}_{t=1}^T,\{\bU_k,\bSig_k\}_{k=1}^K$;
    \Repeat	
    \State \textit{\% update variational parameters}
    \State update $ \balp_t$ by APG;
    \State update $\bU_k$ by APG;
    \myState {$ \bSig_k = \frac{1}{1/\bQ + \bone \diag(\bR_{\bs}^k)/\sigma^2}  $;}
    \myState {$ \omega_t=\frac{\gamma \cdot \pout(\by_t)}{\gamma \cdot \pout(\by_t) + (1-\gamma)\cdot \exp(\hl_{t,k}(\sigma^2,\bphi;\by_t) )} $; }
    \State \textit{\% update model parameters}
    \myState{$\bbA = \textstyle \frac{1}{K}\sum_{k=1}^K \bU_k $;}
    \myState{$\bQ = \textstyle \frac{1}{K}\sum_{k=1}^K (\bU_k-\bbA)^2+\bSig_k$;}
    \myState{$\sigma^2 = \frac{\sum_{k=1}^K  \epsilon_{k} }{M\times (\sum_{t=1}^T\bomg_t)}$;}
    \myState  {$\gamma = \frac{1}{T}\sum_{i=1}^T \omega_t$; }
    \Until {a stopping rule is satisfied}
    \State \textit{\% calculate posterior mean of $\bA_k$ and $\bs_t$}
    \State $\hbA_k = \bU_k, \hbs_t = \frac{\balp_t}{\bone^\top\balp_t} $ ;
    \vspace{0.3em}
    \Ensure $\{ \hbA_k\}_{k=1}^K, \{\hbs_t,\omega_t\}_{t=1}^T$.
\end{algorithmic}
\end{algorithm}

\begin{table*}[t]\centering
\caption{Supported prior distributions $p_{\btheta_A}(\bA_k)$ and  variational posteriors in our framework.}
\vspace{-0.5em}
\label{tab:support dist}
\resizebox{\textwidth}{!}{
\begin{tabular}{|c|c|c|c|c|}\hline
 Prior $p_{\btheta_A}(\bA_k)$ & Support & $q_k(\bA_k;\bphi_k^A)$ & Moments & Reason  \\ \hline\hline
 
\shortstack {Beta \\~\\ $\Beta(\bA_k;\bC,\bD) $}  &
$[0,1]$    & 
\shortstack {Beta \\ ~\\ $\Beta(\bA_k;\bU_k,\bV_k) $ } & 
$\begin{aligned}
\Exp[\bA_k] &= \frac{\bU_k}{\bU_k+\bV_k}\\
\Exp[\bA_k^\top \!\bA_k] &= \Exp[\bA_k]^\top\Exp[\bA_k] + \Diag(\Exp[\bA_k]^\top \tbV_k )  \\
D_\kl( q_k \| p) &=  \log \!B(\bC,\bD)  - \innerF {\bC\!-\!\bU_k}{\psi(\bU_k)} \\
    &\quad~ - \!\log \!B(\bU_k,\bV_k) - \innerF{\bD\!-\!\bV_k}{\psi(\bV_k) } \\
    &\quad~ + \innerF{\bC\!+\!\bD\!-\!\bU_k\!-\!\bV_k} {\psi(\bU_k+\bV_k)}\\
{\rm where ~} \tbV &= \frac{\bV_k}{ (\bU_k+\bV_k)\odot(\bU_k+\bV_k+1)}    
\end{aligned} 
$
& bounded support \\ \hline

\shortstack {Gaussian \\~~\\ $\setN(\bA_k;\bbA,\bQ) $ } & $(-\infty,\infty)$ & 
\shortstack {Gaussian \\~~\\ $\setN(\bA_k;\bU_k,\bSig_k)$ } & 
$\begin{aligned}
\Exp[\bA_k] &= \bU_k\\
\Exp[\bA_k^\top \!\bA_k] &= \bU_k^\top\bU_k + \Diag(\bone^\top\bSig_k) \\
D_\kl( q_k \| p) &= \frac{1}{2}\!\!\sum_{m=\!1}^M \!\sum_{n=1}^N \!\!\frac{[\bU\!_k \!\!-\! \bbA]_{mn}^2 \!+\! [\bSig_k]_{mn} }{ \bQ_{mn} } \!\!+\! \log \!\frac{\bQ_{mn}}{[\bSig_k]_{mn}}\!-\!1
\end{aligned}$ 
& closed-form update \\ \hline

\shortstack {~~\\Log-normal \\~~\\ $\lognormal(\bA_k;\bbA,\bQ)$ } & $(0,\infty)$   &
\shortstack {~~\\Log-normal \\~~\\ $\lognormal(\bA_k;\bU_k,\bSig_k)$ } &  
$\begin{aligned}
\Exp[\bA_k] &= \exp(\bU_k+\bSig_k/2)\\
\Exp[\bA_k^\top \!\bA_k] &=  \Exp[\bA_k]^\top\Exp[\bA_k] + \Diag(\bone^\top \bV_k) \\
D_\kl( q_k \| p) &=  \frac{1}{2}\!\!\sum_{m=\!1}^M \!\sum_{n=1}^N \!\!\frac{[\bU\!_k \!\!-\! \bbA]_{mn}^2 \!+\! [\bSig_k]_{mn} }{ \bQ_{mn} } \!\!+\! \log \!\frac{\bQ_{mn}}{[\bSig_k]_{mn}} \!-\!1\\
{\rm where ~} \bV_k &= (\exp(\bSig_k)-\bone)\odot \exp(2\bU_k+\bSig_k)
\end{aligned}$ 
& nonnegative support \\ \hline

\shortstack {~~\\ Gamma \\~~\\  Gamma$(\bA_k;\bC,\bD) $} &
$(0,\infty) $    &
\shortstack {~~\\Gamma \\~~\\ Gamma$(\bA_k;\bU_k,\bV_k)$} & 
$\begin{aligned}
\Exp[\bA_k]  &= \frac{\bU_k}{\bV_k}\\
\Exp[\bA_k^\top \!\bA_k] &=  \Exp[\bA_k]^\top\Exp[\bA_k] + \Diag(\bone^\top \tbV_k), ~~\tbV_k=\frac{\bU_k}{\bV_k^2} \\
D_\kl( q_k \| p) &= \innerF {\bC-\bU_k} {\psi(\bC)} - \log \Gamma(\bC) + \log\Gamma(\bU_k)\\
&\quad ~+ \innerF {\bU_k} {\log \bD-\log\bV_k} + \innerF{\bC}{\frac{\bV_k}{\bD}-\bone}
\end{aligned}$ 
& nonnegative support \\ \hline

\shortstack {~~\\ Uniform \\~~\\  $\setU_{[\bzero,\bone]}(\bA_k) $} & $[0,1]$ &
\shortstack {~~\\Beta \\~~\\  $\Beta(\bA_k;\bU_k,\bV_k) $ } & 
$\begin{aligned}
\Exp[\bA_k] &= \frac{\bU_k}{\bU_k+\bV_k}\\
\Exp[\bA_k^\top \!\bA_k] &= \Exp[\bA_k]^\top\Exp[\bA_k] + \Diag(\Exp[\bA_k]^\top \tbV_k )  \\
D_\kl( q_k \| p) &= -\log B(\bU_k,\bV_k)  - \innerF{\bU_k \!+ \!\bV_k\!-\!2} {\psi(\bU_k+\bV_k)} \\
&\quad ~ + \innerF{\bU_k-\bone} {\psi(\bU_k)} + \innerF{\bV_k-\bone} {\psi(\bV_k)}
\end{aligned}$  
& bounded support \\ \hline

\multicolumn{3}{|c}{
 $\begin{aligned}
\Beta(\bX;\bC,\bD) & := \prod_{m,n}\frac{1}{B(C_{mn},D_{mn})} X_{mn}^{C_{mn}-1} (1-X_{mn})^{D_{mn}-1} \\
\setN(\bX;\bU,\bQ) & := \prod_{m,n} \setN(X_{mn};U_{mn},Q_{mn})
\end{aligned}$ 
} &
\multicolumn{2}{c|}{
 $\begin{aligned}
\lognormal(\bX;\bU,\bQ) & := \prod_{m,n} \frac{1}{X_{mn}\sqrt{2\pi Q_{mn}}} \exp \left( -\frac{(\ln X_{mn}-U_{mn})^2} {2Q_{mn}} \right)\\
{\rm Gamma}(\bX;\bC,\bD) & := \prod_{m,n} \frac{D_{mn}^{C_{mn}}}{\Gamma(C_{mn})} X_{mn}^{C_{mn}-1} \exp(-D_{mn} X_{mn}) \\
\setU_{[\bzero,\bone]}(\bX) &= \prod_{m,n}\indfn_{[0,1]}(X_{mn})
\end{aligned}$ 
} 
  \\ \hline 

\end{tabular}}
\end{table*}

\section{Numerical Results}
\label{sec:exp}
In this section, we use synthetic, semi-real, and real data experiments to validate our algorithm design.

\subsection{General Settings}
\subsubsection{Baselines} 
The baselines include the MF-type algorithm \texttt{ELMM} \cite{drumetz2016blind}, the library-based method \texttt{MESMA} \cite{roberts1998mapping}, the Bayesian inference algorithm \texttt{UsGNCM} \cite{halimi2015unsupervised}, the EM algorithm \texttt{K-Gaussians} \cite{woodbridge2019unmixing}, and the outlier-robust MF-type
algorithm \texttt{VOIMU} \cite{syu2019outlier}. 
Besides these HU-EV algorithms, we also include the LMM-based, outlier-robust \texttt{SISAL} algorithm \cite{Dias2009} as an additional benchmark.

\subsubsection{Algorithm Settings}
We follow the default parameter settings as suggested by the baseline papers unless otherwise specified.
For the regularization parameters, we set $\tau=10^{-3}$ for \texttt{SISAL}, $(\lambda_S,\lambda_A,\lambda_\Psi) =(0.5, 0.015,0.05)$ for \texttt{ELMM}, and $\lambda_1=\lambda_2=1$ for \texttt{VOIMU}. We use $p=0.5$ in \texttt{VOIMU}.
The library for \texttt{MESMA} is constructed following \cite{borsoi2021spectral}.
For initialization, we use \texttt{SISAL}-estimated endmembers for 
all the HU-EV methods. 
When the algorithm needs abundance initialization (e.g., ELMM, \texttt{K-Gaussians} and \texttt{VOIMU}), we feed them the \texttt{FCLS}-output \cite{heinz2001fully} abundances using the \texttt{SISAL}-estimated endmembers.
The \texttt{K-Gaussians} and \texttt{VOIMU} algorithms are stopped if the relative change of objective function is less than $10^{-4}$ or when the number of iterations exceeds 100 and 500,  respectively.
These parameters are tuned for their best performance according to our empirical observations.
For \texttt{HELEN}, since the objective function is not directly calculated, we stop the algorithm if the relative change of the prior mean of $\bA_k$ is less than $10^{-5}$ or when 300 iterations are executed. 
We set the outlier distribution in \texttt{HELEN} as the zero-mean Gaussian distribution with variance $\sigma^2=3^2$, in order to capture a wide range of outlier magnitudes.

\begin{figure*}[t!]
\vspace{-2.5em}
\centering
\hspace{-3.5em}
\includegraphics[width=1.06\linewidth]{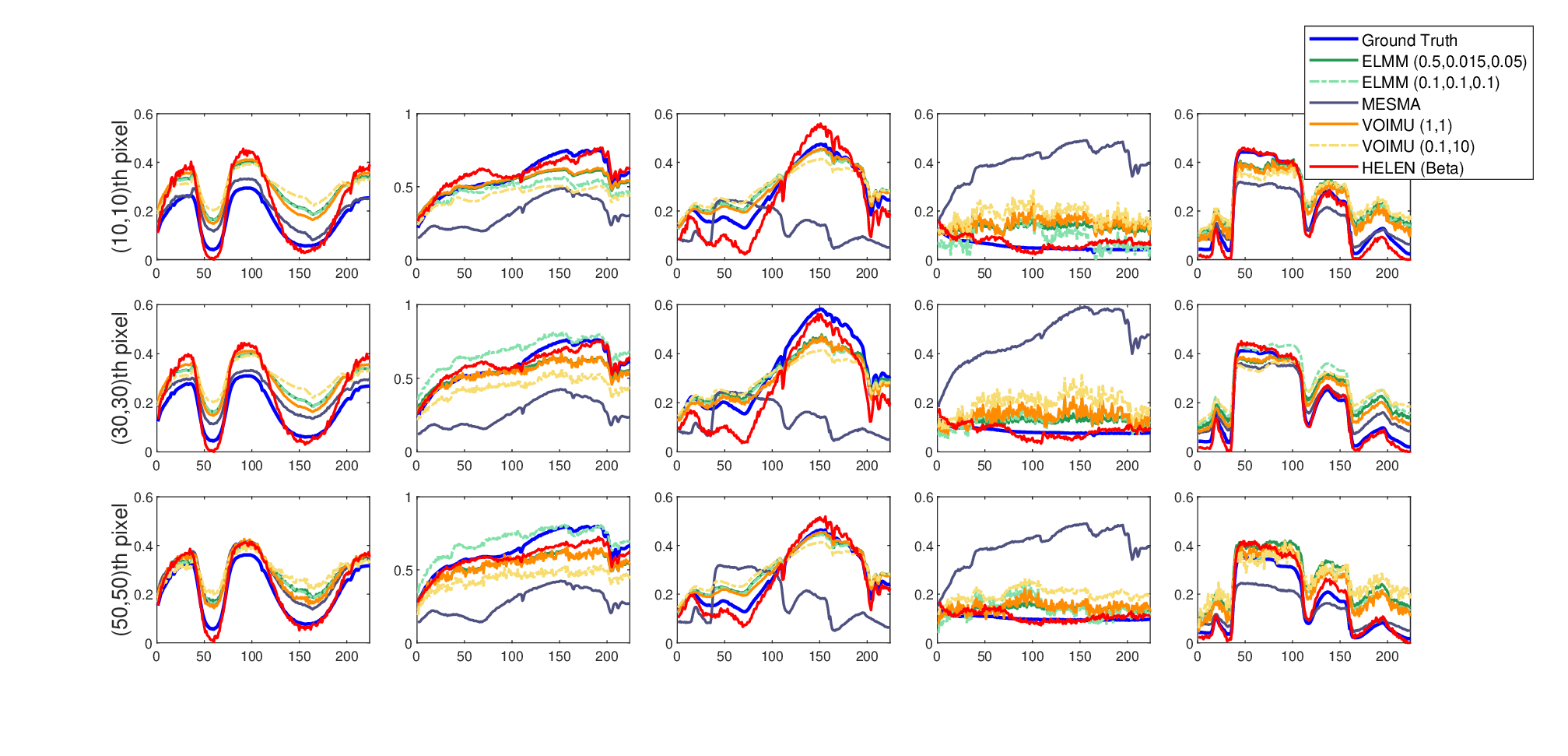}
\vspace{-4.5em}
\caption{(Synthetic data experiment) The ground-truth and the algorithm-estimated $\bA_t$ in three selected pixels located at $(10,10), (30,30) $ and $(50,50)$.}
\label{fig:syn A}
\end{figure*}

\begin{figure*}[t!]
\hspace{-2em}\centering
\includegraphics[width=0.94\linewidth]{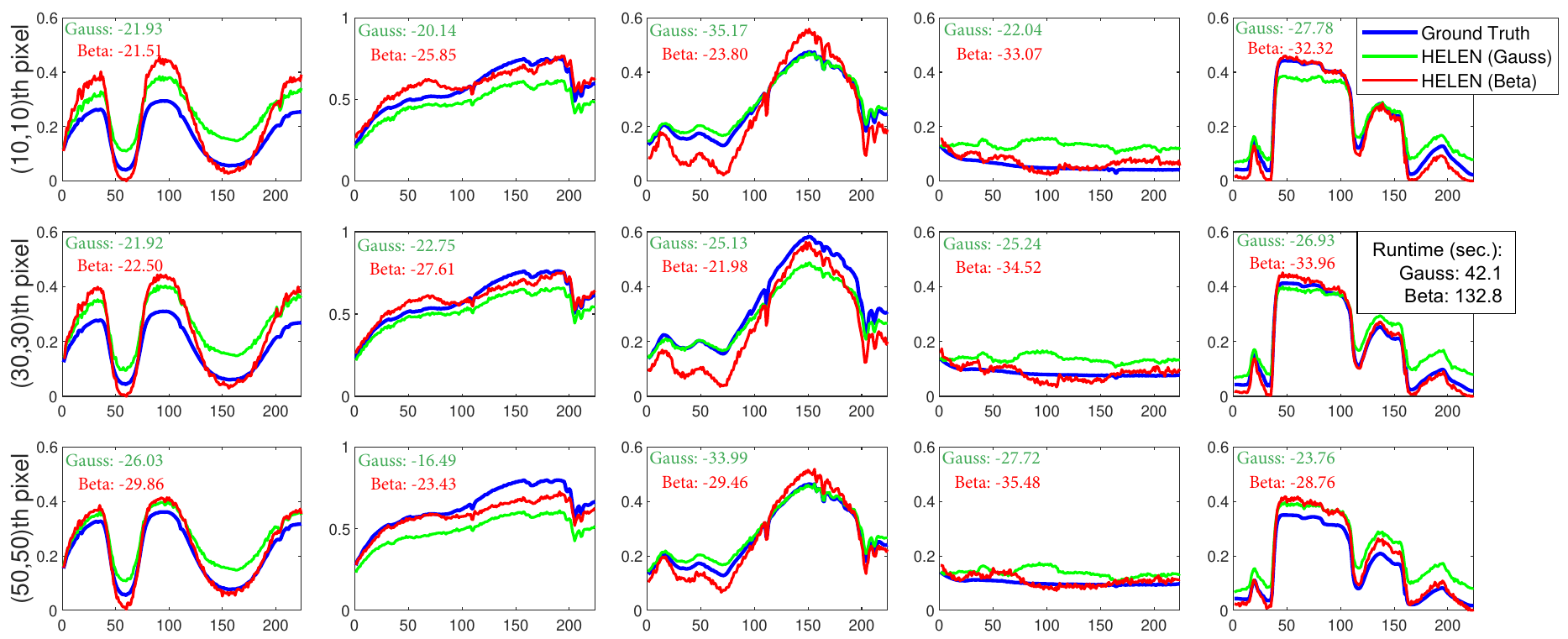}
\caption{(Synthetic data experiment) The ground-truth and the \texttt{HELEN}-estimated $\bA_t$ in three selected pixels located at $(10,10), (30,30) $ and $(50,50)$. The MSEs of \texttt{HELEN} that use Gaussian and Beta prior are marked in green and red. The runtime is also shown.}
\label{fig:syn At BECAprior}
\end{figure*}

\subsubsection{Metrics}
We use three metrics, namely, the average {\it spectral angle mapper} (SAM) and {\it mean square error} (MSE) of the estimated $\hat{\bm A}_t$'s, and the average {\it root mean square error} (RMSE) of the estimated $\hbs_t$'s, \ie
\begin{align*}
    {\rm SAM}_\bA &= \frac{1}{NT}\sum_{t=1}^T \sum_{n=1}^N \cos^{-1}\left( \frac{\ba_{nt}^\top \hba_{nt}}{\|\ba_{nt}\|_2 \| \hba_{nt}\|_2 } \right), \\
    {\rm MSE}_\bA &= \frac{1}{NT}\sum_{t=1}^T \| \bA_t-\hbA_t \|_F^2  , \\
    {\rm RMSE}_\bs &=  \frac{1}{T}\sum_{t=1}^T \sqrt{ \| \bs_t - \hbs_t \|_2^2/N },
\end{align*}
where 
$\hbA_t=[\hba_{1t},\dots,\hba_{Nt}]$ and $\hbs_t$ are the estimated endmembers of the $t$th pixel. Note that we have $\hbA_{t}=\hbA_{t'}$ for all $t,t'\in {\cal I}_k$ due to our algorithm design. However, the ground-truth $\bA_t$ and $\bA_{t'}$ for $t,t'\in {\cal I}_k$ need not be identical in the experiments. 
For the algorithms that do not estimate local endmembers $\bm A_t$ or $\bm A_k$ (i.e., \texttt{SISAL}, \texttt{UsGNCM} and \texttt{K-Gaussians}), we only evaluate their estimated $\hbs_t$'s.

\begin{table*}[th!]    
    \caption{The performance of the schemes on synthetic data (with the best marked in red).}
    \label{tab:syn exp}
    \centering
    \vspace{-0.5em}
    \resizebox{\textwidth}{!}{
    \begin{tabular}{|c|c||c|c|c|c|c|c|c|c|}
        \hline
        \multicolumn{2}{|c||}{Algorithm} &
        \texttt{SISAL}+\texttt{FCLS} &
        \texttt{ELMM} &
        \texttt{MESMA} &
        \texttt{UsGNCM} &
        \texttt{K-Gaussians} &
        \texttt{VOIMU} &
        \texttt{HELEN (Beta)} &
        \texttt{HELEN (Gauss)}  \\ \hline\hline
         
        \multicolumn{1}{|c|}{\multirow{3}{*}{no outlier}} &
        SAM$_\bA \downarrow$ & -  & 4.40 & 5.39 & - & - & 4.30    &  3.78   & \red 3.22 \\ 
        \multicolumn{1}{|c|}{}  &
        MSE$_\bA \downarrow$ & - & -18.51 & -15.43 & - & - & -18.70   &  \red -21.46  & -20.58 \\ 
        \multicolumn{1}{|c|}{}  & 
        RMSE$_\bs \downarrow$& 0.1391  & 0.1439 & 0.1790 & 0.1044 & 0.1254 & 0.1718  &  \red 0.0862  & 0.0973 \\ \hline

        \multicolumn{1}{|c|}{\multirow{3}{*}{100 outliers}} &
        SAM$_\bA \downarrow$ & -  & 4.74 & 9.16 & - & - & 4.54   &  3.36  & \red 3.27 \\ 
        \multicolumn{1}{|c|}{}  &
        MSE$_\bA \downarrow$ & - & -17.39 & -7.44 & - & - & -17.92   & \red -21.30   & -20.79 \\ 
        \multicolumn{1}{|c|}{}  & 
        RMSE$_\bs \downarrow$& 0.1518  & 0.1635 & 0.2497 & 0.1428 & 0.1386 & 0.1823   & \red 0.0910  & 0.0990 \\ \hline
       
    \end{tabular}}
    \vspace{-1em}
\end{table*}

\begin{table}[th!]    
    \caption{The performance of \texttt{HELEN (Beta)} with different block size on synthetic data.}
    \label{tab:syn block size}
    \centering
    \resizebox{\linewidth}{!}{
    \begin{tabular}{|c|c||c|c|c|c|c|c|}
        \hline
        \multicolumn{2}{|c||}{\multirow{2}{*}{Algorithm}} &
        \multicolumn{3}{c|}{\texttt{HELEN (Beta)}} &
        \multicolumn{3}{c|}{\texttt{HELEN (Gauss)}} \\ \cline{3-8}
        \multicolumn{2}{|c||}{} &   
        $4\times 4$ &
        $5\times 5$ &
        $\!\! 10\times 10\!\!$  &
        $4\times 4$ &
        $5\times 5$ &
          $\!\!10\times 10\!\!$  \\ \hline\hline
         
        {\multirow{3}{*}{no outlier}} &
        SAM$_\bA \downarrow$  &3.99  &  3.78 & 3.40  & \red 3.18  &   3.22 & 3.27  \\ 
        & MSE$_\bA \downarrow$   & -20.80 & \red -21.46  &  -21.33  & -20.36 & -20.58  &  -20.14 \\
        & RMSE$_\bs \downarrow$   &  0.0903  & \red 0.0862  & 0.0865 &  0.0998  &  0.0973  & 0.1023 \\ \hline

        {\multirow{3}{*}{\!\!100 outliers\!\!}} &
        SAM$_\bA \downarrow$ & 3.52  &  3.36 &  3.33  & 3.34  &   \red 3.27 & 3.42 \\ 
        &
        MSE$_\bA \downarrow$  & -20.85 & \red -21.30  &  -20.90  &  -20.09  & -20.79 & -19.99 \\ 
        & 
        RMSE$_\bs \downarrow$ &  0.0938  & \red 0.0910  & 0.0918 &  0.1063  &  0.0990  & 0.1081\\ \hline
    \end{tabular}}
\end{table}

\begin{figure}[tb]
    \centering
    \includegraphics[width=0.95\linewidth]{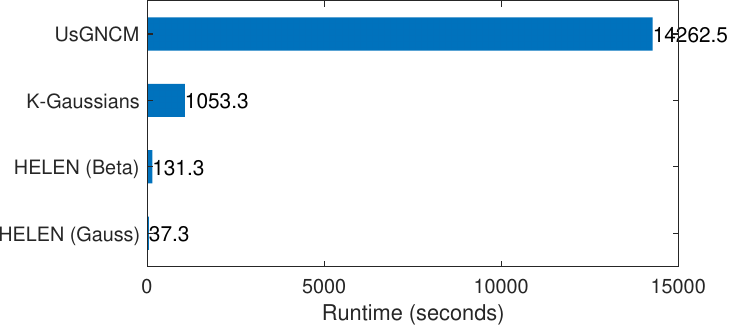}
    \caption{Runtime of probabilistic schemes on synthetic data. The settings are the same as those in Table~\ref{tab:syn exp} (no outliers).}\vspace{-1em}
    \label{fig:syn runtime}
\end{figure}

\subsection{Synthetic Data Experiment}
\subsubsection{Data Generation}
The synthetic images were generated following the block model in \eqref{eq:ourmodel} with $5\times 5$ non-overlapping blocks. 
The total size of the image is chosen as $100\times 100$. We generate $\bs_t$'s from the uniform Dirichlet distribution and remove the $\bs_t$'s such that $\|\bs_t\|_\infty > 0.7$. This step removes pixels close to pure endmembers, making the unmixing problem more challenging \cite{Dias2009}.
To synthesize the EV, in each trial, we randomly select five materials 
from the USGS library \cite{clark2007usgs} (with $M=224$) and generate 25 variants for each spectrum using the physical models from \cite{borsoi2021spectral}.  
Specifically, in each trial we simulate two of the randomly selected materials using a simplified Hapke's model \cite{hapke1981bidirectional}, and another two materials using a simple atmospheric compensation model \cite{ramakrishnan2015shadow}.  
The fifth material's variations are simulated using the PROSPECT-D model \cite{feret2017prospect}.
The five columns of $\bA_k$ are then randomly selected from the five classes of endmembers.
To mimic the gradual changing of spectra in space, we generate similar but non-identical $\bA_t$'s from the $\bA_k$'s for all $t\in {\cal I}_k$. This is done by applying a 2D Gaussian blurring kernel with a size of 11 and a unit standard deviation to $\bA_k$.
Finally, we add the zero-mean i.i.d. Gaussian noise and set the signal-to-noise ratio (SNR) to be 25dB.
When testing the outlier robustness, we additionally replace 100 randomly selected pixels with outliers that follow elementwise uniform-[0,2] distribution; see similar settings in \cite{wu2017stochastic,steinbuss2021benchmarking}. We intentionally make the outlier distribution different from the outlier prior used by \texttt{HELEN} (which is Gaussian), also in order to test our algorithm's robustness under prior mis-specifications.

\subsubsection{Results}
Fig. \ref{fig:syn A} shows the result of an illustrative example.
To clearly see how well the methods cope with EV, we use an outlier-free setting here.
In the figure, we select three pixels, located at $(10,10), (30,30) $ and $(50,50)$, respectively,
to show the ground-truth and the algorithm-estimated $\bA_t$. 
The results of \texttt{SISAL}, \texttt{UsGNCM} and \texttt{K-Gaussians} are not shown in the figure as they do not estimate $\bA_t$. For \texttt{HELEN}, we use $5\times 5$ non-overlapping square blocks as ${\cal I}_k$ for $k=1\ldots,25$.
One can see that \texttt{MESMA} does not perform well in this simulated image, as it uses the so-called pure pixels (i.e., pixel with $\|\bs_t\|_{\infty}=1$) to construct its library, which does not exist under our setting.
The results of the MF-type algorithms, \texttt{ELMM} and \texttt{VOIMU}, attain more reasonable results. 
Nonetheless, the MF-type methods require parameter tuning and they are sensitive to the parameter selection---one can see that the two sets of relatively similar hyper-parameters under test yield rather different estimations.
Compared to the baselines, \texttt{HELEN (Beta)} outputs visually more accurate estimations for $\bA_t$ in the three pixels.

Fig. \ref{fig:syn At BECAprior} further shows the performance of \texttt{HELEN (Beta)} and \texttt{HELEN (Gauss)} under the same settings. The Beta prior generally leads to a visually more appealing estimation for $\bA_t$'s in the displayed pixels. 
Nonetheless, \texttt{HELEN (Gauss)} also occasionally outperforms its Beta counterpart (see the third column on Fig.~\ref{fig:syn At BECAprior}). More importantly, the simple updates make \texttt{HELEN (Gauss)} around 3 times faster than \texttt{HELEN (Beta)} in this case (42.1 seconds v.s. 132.8 seconds).

\begin{figure}[t!]
\centering
\subfigure[color image]{
    \label{fig:semireal img}
    \includegraphics[width=0.4\columnwidth]{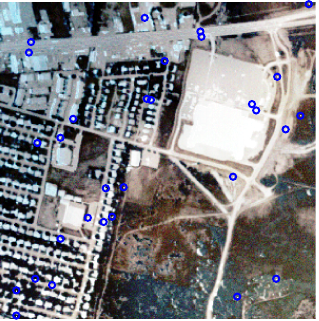}}\quad
\subfigure[\!\texttt{VOIMU}]{
    \label{fig:semireal ouliermap voimu}
    \includegraphics[width=0.4\linewidth]{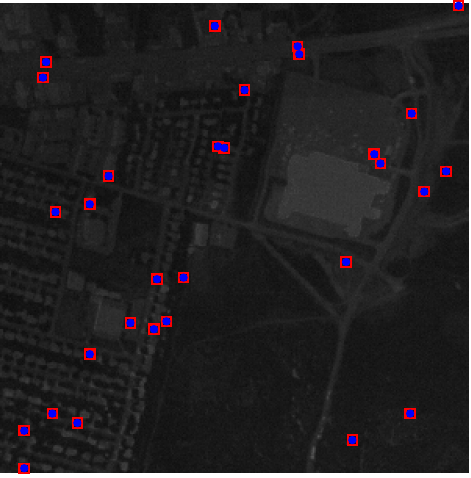}}
\subfigure[\!\!\texttt{HELEN (Beta)}\!]{
    \label{fig:semireal ouliermap rebeca}
    \includegraphics[width=0.4\linewidth]{fig/semiFig/semireal_outliermap.pdf}}\quad
\subfigure[\!\!\texttt{HELEN (Gauss)}\!\!\!\!\!]{
    \label{fig:semireal ouliermap rebeca}
    \includegraphics[width=0.4\linewidth]{fig/semiFig/semireal_outliermap.pdf}}
\caption{(Semi-real data experiment) The generated color image and detected outlier maps (along with the image using the 30th band) by \texttt{VOIMU} and \texttt{HELEN}. The 30 artificially added outliers are marked by blue circles. The detected outliers by the algorithms are marked by red squares.}\vspace{-1em}
\end{figure}

Table \ref{tab:syn exp} shows the average SAM, MSE and RMSE of the algorithms using the synthetic data under both outlier-free and outlier-present cases. The results of all the algorithms are averaged from 100 Monte Carlo trials, except for \texttt{UsGNCM}---whose results are averaged from 10 trials due to its prohibitive computational cost. 
One can see that in the outlier-free case, the probabilistic HU-EV algorithms, namely \texttt{UsGNCM} and \texttt{HELEN}, yield the similar low RMSE$_\bs$'s. 
However, \texttt{UsGNCM} does not provide $\bA_t$'s and its runtime is around 100 times and 400 times slower than \texttt{HELEN (Beta)} and \texttt{HELEN (Gauss)},  respectively---see Fig. \ref{fig:syn runtime}.
In the case with outliers,  the \texttt{HELEN}-based methods show promising performance, outperforming the best-performing baseline by nontrivial margins in all metrics.

Table \ref{tab:syn block size} shows the performance of \texttt{HELEN} with three different block sizes, namely $4\times 4, 5\times 5$ and $10\times 10$. 
Note that the settings are identical to those in Table~\ref{tab:syn exp}, where
the ground-truth block size is $5\times 5$.
One can see that \texttt{HELEN} with different block sizes yield comparable estimation results in all of the metrics, indicating reasonable robustness to block size mis-specifications.
In the sequel, we will use $10 \times 10 $ non-overlapping square blocks as our default setting for \texttt{HELEN}.

\begin{figure}[t]
    \vspace{-1em}
    \centering
    \!\!\!\!\!\includegraphics[width=1.1\linewidth]{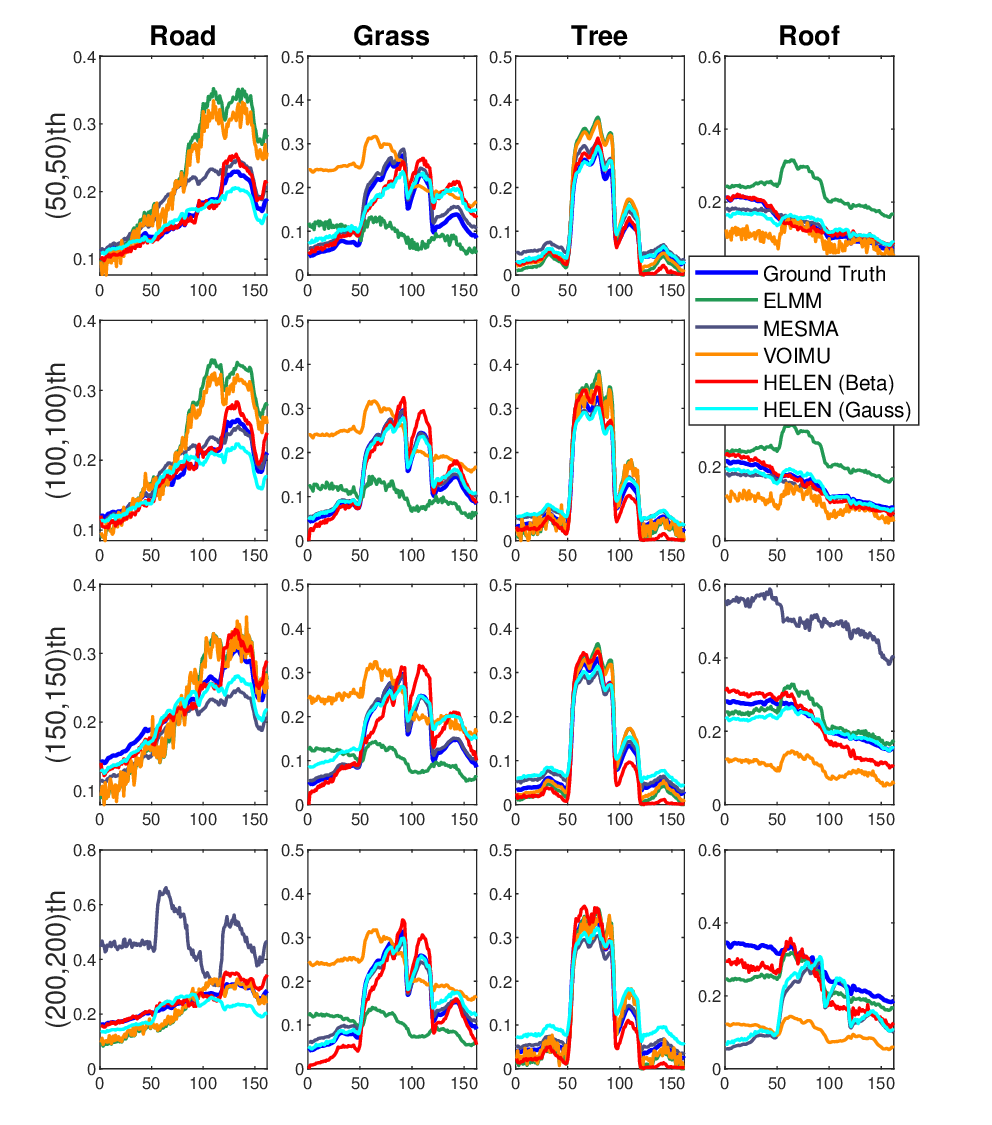}
    \vspace{-2.5em}
    \caption{(Semi-real data experiment) The ground truth and estimated $\bA_t$ in four selected pixels, located at $(50,50),(100,100),(150,150)$ and $(200,200)$. }
    \label{fig:semireal At}
\end{figure}

\begin{figure}[t]
\vspace{-1em}
    \centering
    \includegraphics[width=1\linewidth]{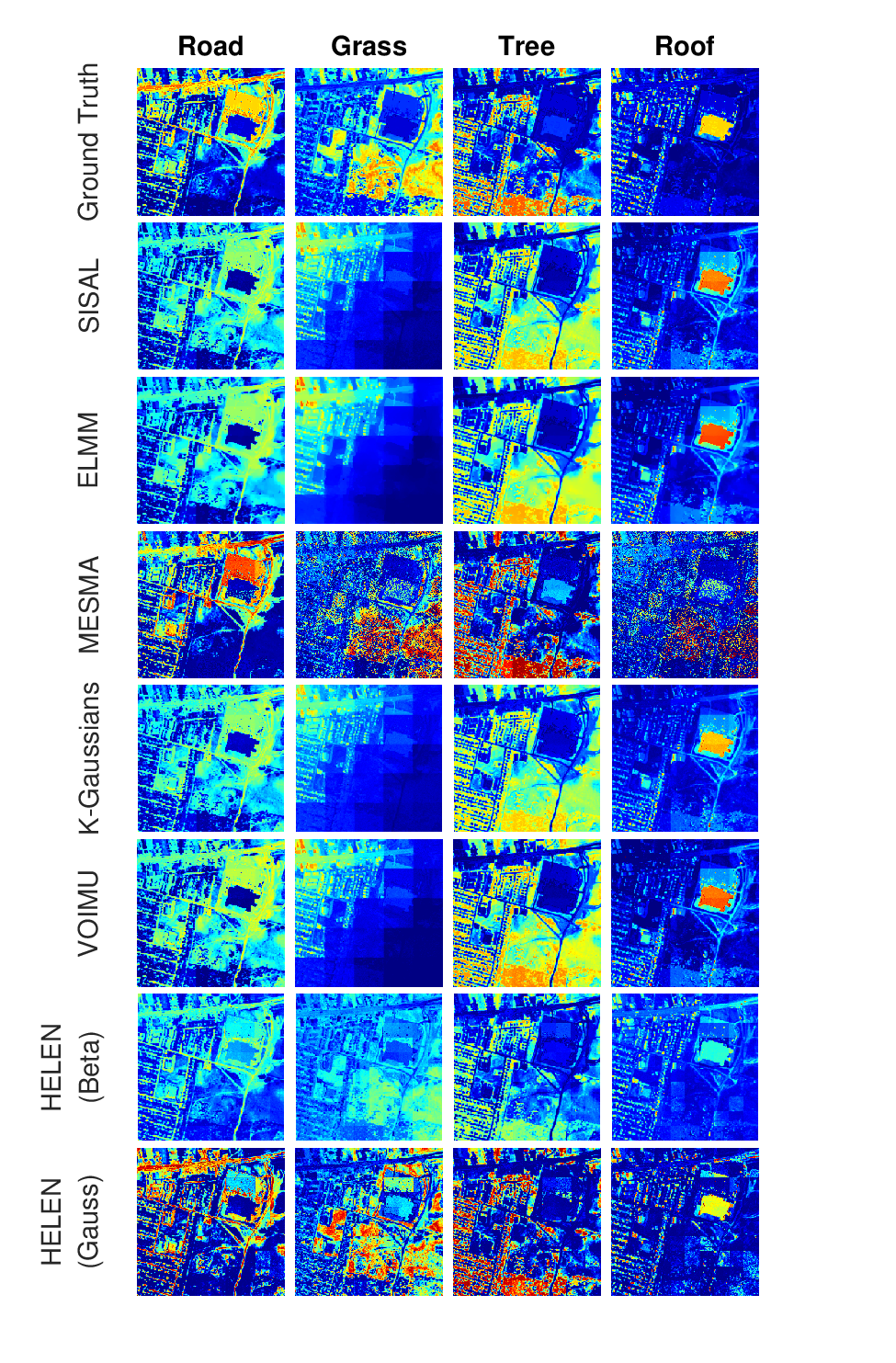}
    \vspace{-3.5em}
    \caption{(Semi-real data experiment) Ground truth and estimated abundance maps by the algorithms.}
    \vspace{-1em}
    \label{fig:semireal S}
\end{figure}

\begin{table}[t]    
    \caption{The performance of the schemes on semi-real data (with the best and second best marked in red and blue, respectively).}
    \label{tab:semi exp}
    \centering
    \vspace{-0.5em}
    \resizebox{\linewidth}{!}{\Huge
    \begin{tabular}{|c||c|c|c|c|c|c|c|}
        \hline
        Algorithm  & \texttt{SISAL} & \texttt{ELMM} & \texttt{MESMA} & \texttt{K-Gaussians} & \texttt{VOIMU} & \texttt{HELEN (Beta)} & \texttt{HELEN (Gauss)} \\ \hline \hline 
        SAM$_\bA \downarrow$  & -  & 6.21  & 5.43 & - & 6.09 & \red 4.09   & \blue 4.14 \\ 
        MSE$_\bA \downarrow$ &- & -23.34 & -20.81 & - & -24.14 & \red  -29.55 & \blue -28.01  \\ 
        RMSE$_\bs \downarrow$ & 0.1950  & 0.1996 & 0.1425 & 0.1871 & 0.2001    & \blue 0.1076 & \red 0.1052 \\ \hline
        \!Time (sec.)\! $\downarrow$\! &\red  16.8  & 787.0 & 158.1 & 3248.2 & 502.8  & 354.3   & \blue  133.7  \\ \hline
    \end{tabular}}
\end{table}

\subsection{Semi-Real Data Experiment}

\subsubsection{Data Generation}
In this subsection, we use the pre-processed Urban data, in which the ``ground-truth'' endmembers and abundances are provided \cite{zhu2017hyperspectral}.
The Urban image has 162 bands and 4 materials, namely, ``road'', ``grass'', ``tree'', and ``roof''.
We use a $300\times 300$ subarea and partition it into $5\times 5$ blocks. 
In each block, we generate $\bba_{i,k}$ by the average of the pixels whose ground-truth abundance vectors satisfy $s_{i,t} \geq 0.9$ (which means $\by_t\approx \ba_{i,t}$) for $t\in {\cal I}_k$.
For the abundance maps, we blur the ground truth using a Gaussian kernel with a standard deviation of 0.5 and kernel size of 3.
Following the settings before, we truncate the elements of $\bm s_t$ that are larger than 0.8.
This procedure retains the authenticity of the abundance maps and removes approximate pure pixels (i.e., the pixels with $\|\bs_t\|_{\infty}\approx 1$)---making the unmixing problem more challenging. 
The SNR is set to be 25dB.
Finally, 30 pixels are replaced by outliers that are generated as before. 
The constructed hyperspectral image is shown in
Fig.~\ref{fig:semireal img} (only the RGB bands are displayed), where the outliers are marked by the blue circles.

\begin{table*}[ht]
\centering
\caption{The MSE performance on real data. }
\label{tab:real exp}
\begin{tabular}{|c|ccc|ccc|ccc|}
\hline
\multirow{2}{*}{\!\!Algorithm\!\!} &
  \multicolumn{3}{c|}{water} &
  \multicolumn{3}{c|}{rock} &
  \multicolumn{3}{c|}{vegetation} \\ \cline{2-10} 
 &
  \multicolumn{1}{c|}{(4,5)} &
  \multicolumn{1}{c|}{(5,6)} &
  (4,8) &
  \multicolumn{1}{c|}{(1,9)} &
  \multicolumn{1}{c|}{(5,1)} &
  (8.10) &
  \multicolumn{1}{c|}{(8,1)} &
  \multicolumn{1}{c|}{(10,1)} &
  (7,3) \\ \hline\hline
\texttt{ELMM} &
  \multicolumn{1}{c|}{-34.59} &
  \multicolumn{1}{c|}{-30.93} &
   -41.13 &
  \multicolumn{1}{c|}{-18.43} &
  \multicolumn{1}{c|}{-18.89} &
  -26.06 &
  \multicolumn{1}{c|}{\red -36.14} &
  \multicolumn{1}{c|}{-24.90} &
  \red -29.84 \\ 
\texttt{MESMA} &
  \multicolumn{1}{c|}{-22.11} &
  \multicolumn{1}{c|}{-24.53} &
  \red -41.50 &
  \multicolumn{1}{c|}{-19.67} &
  \multicolumn{1}{c|}{-22.95} &
  -27.43 &
  \multicolumn{1}{c|}{-31.89} &
  \multicolumn{1}{c|}{\red -27.02} &
  -25.16 \\ 
\texttt{VOIMU} &
  \multicolumn{1}{c|}{-33.69} &
  \multicolumn{1}{c|}{-27.97} &
  -40.48 &
  \multicolumn{1}{c|}{-18.25} &
  \multicolumn{1}{c|}{-20.39} &
  -23.90 &
  \multicolumn{1}{c|}{\blue -31.94} &
  \multicolumn{1}{c|}{-25.26} &
  \blue -29.00 \\ 
\rowcolor{blue!5} \texttt{HELEN (Beta)} &
  \multicolumn{1}{c|}{\blue -37.10} &
  \multicolumn{1}{c|}{\red -38.24} &
  \blue -41.33 &
  \multicolumn{1}{c|}{\blue -30.36} &
  \multicolumn{1}{c|}{\red -41.79} &
  \blue -38.57 &
  \multicolumn{1}{c|}{-28.42} &
  \multicolumn{1}{c|}{-25.86} &
  -27.14 \\ 
\rowcolor{blue!5}\texttt{HELEN (Gauss)} &
  \multicolumn{1}{c|}{\red -37.72} &
  \multicolumn{1}{c|}{\blue -37.99} &
   -39.02 &
  \multicolumn{1}{c|}{\red -30.56} &
  \multicolumn{1}{c|}{\blue -40.49} &
  \red -38.64 &
  \multicolumn{1}{c|}{-28.98} &
  \multicolumn{1}{c|}{\blue -26.28} &
  -27.51 \\ \hline
\end{tabular}
\end{table*}

\subsubsection{Results} 
Fig. \ref{fig:semireal ouliermap voimu}--\ref{fig:semireal ouliermap rebeca}
show the detected outliers by the outlier-robust HU-EV algorithms, \ie \texttt{VOIMU} and \texttt{HELEN}.
One can see that all the outliers are successfully identified by both algorithms. 
However, the endmembers and abundances estimated by \texttt{VOIMU} are less accurate, as shown in
Fig. \ref{fig:semireal At}. There, the ground-truth and estimated $\bA_t$ in four selected pixels, located at $(50,50)$, $(100,100)$, $(150,150)$ and $(200,200)$, respectively, 
are displayed.
One can see that all the baselines, including \texttt{VOIMU}, appear to be negatively affected by the existence of the outliers.
Nonetheless, the outlier-induced impact is alleviated to a good extent in the results of \texttt{HELEN (Beta)} and \texttt{HELEN (Gauss)}. 
Similar results are seen in
Fig. \ref{fig:semireal S} that shows the estimated abundance maps. That is, \texttt{HELEN (Beta)} and \texttt{HELEN (Gauss)} generally output visually more accurate abundance maps relative to others.

Table \ref{tab:semi exp} shows the quantitative evaluation results. One can see that \texttt{HELEN}'s MSE$_{\bm A}$ is around 5dB lower than that of the best-performing baseline (\ie \texttt{VOIMU}), which is a significant margin.
The SAM$_{\bA}$ and RMSE$_{\bs}$ values obtained by \texttt{HELEN} are also visibly improved upon those of the baselines.

\begin{figure}[t]
\centering
\subfigure[color image]{
    \label{fig:real img}    \includegraphics[width=0.44\columnwidth]{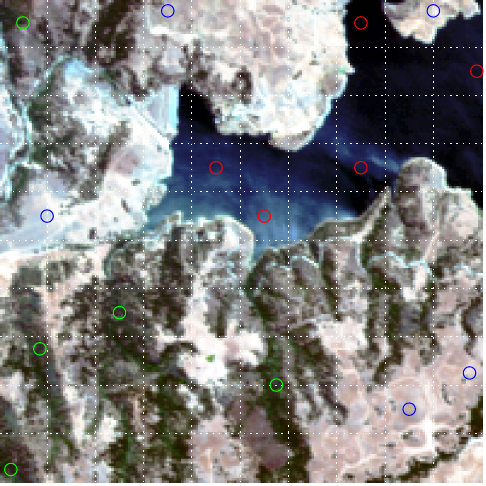} }
\subfigure[reference spectra]{
    \label{fig:real A}
    \includegraphics[width=0.49\linewidth]{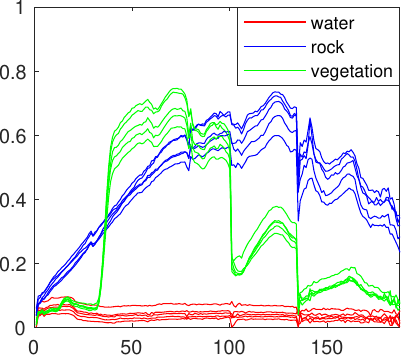} }
\caption{(Real data experiment) Color image of a subarea of Moffett Field (segmented as $10 \times 10 $ blocks) and five manually selected reference spectra for each of the three materials.  }
\end{figure}

\begin{figure}[t]
    \centering
    \subfigure[water]{
    \label{fig:real water}
    \!\!\!\!\!\!\!\!\!\!\!\!\includegraphics[width=1.15\linewidth]{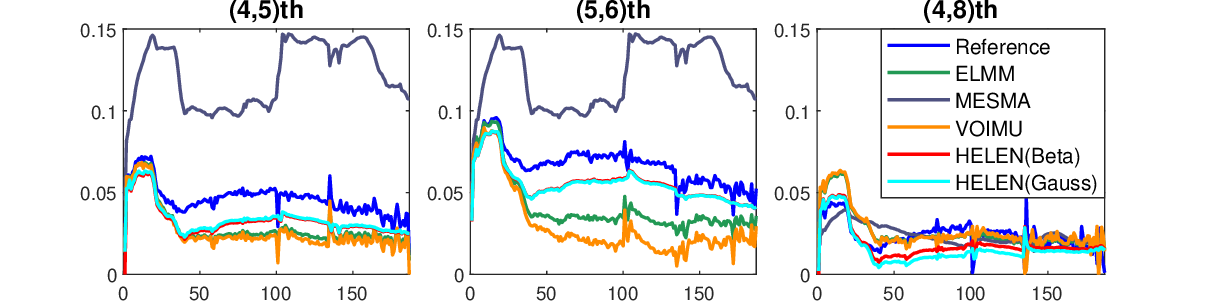} }
    \subfigure[rock]{
    \label{fig:real rock}
    \!\!\!\!\!\!\!\!\!\!\!\!\includegraphics[width=1.15\linewidth]{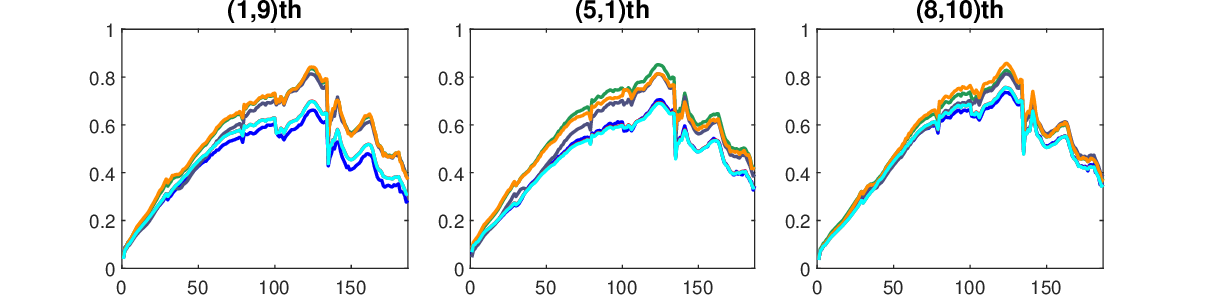}
    }
    \subfigure[vegetation]{
    \label{fig:real veg}
    \!\!\!\!\!\!\!\!\!\!\!\!\includegraphics[width=1.15\linewidth]{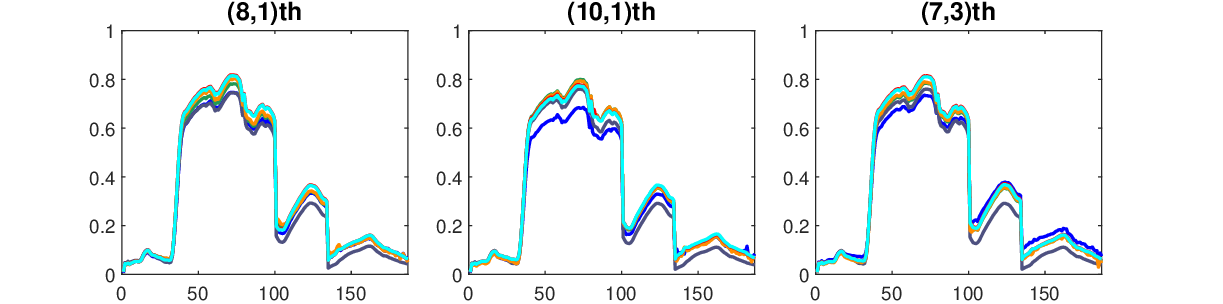}
    }
    \vspace{-1em}
    \caption{(Real data experiment) The reference and estimated local endmembers in 3 selected blocks for water, rock and vegetation.}
    \label{fig:real Ak}
\end{figure}

\begin{figure}[t!]
\centering 
\subfigure[\texttt{VOIMU}]{\!\!\!\!
    \label{fig:real outliermap voimu}
    \includegraphics[width=0.32\columnwidth]{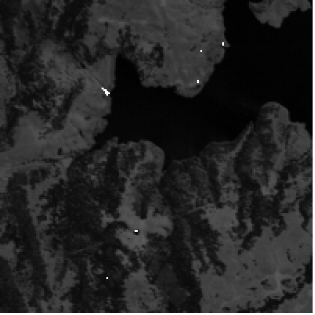}} 
\subfigure[\texttt{HELEN (Beta)}]{\!
    \label{fig:real outliermap beta}
    \includegraphics[width=0.32\columnwidth]{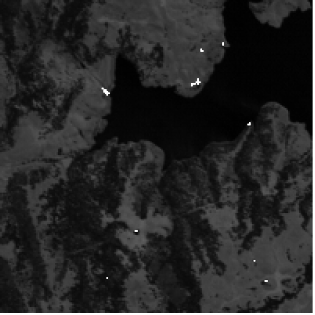}}
\subfigure[\texttt{HELEN (Gauss)}]{\!
    \label{fig:real outliermap gau}
    \includegraphics[width=0.32\columnwidth]{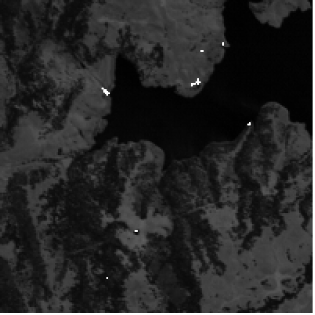}}
\caption{(Real data experiment) The estimated outlier maps by \texttt{VOIMU}, \texttt{HELEN (Beta)} and \texttt{HELEN (Gauss)}. The identified outlier pixels are displayed in white, along with the image using the 170th band.}
\label{fig:real outlier}
\end{figure}

\subsection{Real Data Experiment}
\subsubsection{Dataset}
We use a $200\times 200$ subarea of the Moffett Field data. The corresponding RGB color image is shown in Fig. \ref{fig:real img}, where the $10\times 10$ blocks used by \texttt{HELEN} are also plotted. 
The image is known to contain three prominent materials, namely, rock, vegetation, and water \cite{moffett}. As there is no ground truth available,  for each material we manually select five pixels where likely only one material is present as reference. The selected pixels are marked by colored circles in Fig. \ref{fig:real img}, whose spectra are plotted in Fig. \ref{fig:real A}.

\subsubsection{Results}
Fig. \ref{fig:real Ak} shows the estimated endmembers of the water, rock and vegetation, respectively, in the three selected blocks. 
One can see that \texttt{HELEN} captures variations in both the magnitude and shape of the materials across these blocks. 
Table \ref{tab:real exp} shows the MSE evaluated in the selected blocks (with respect to the reference ``ground-truth'' endmembers). It is seen that \texttt{HELEN (Beta)} and \texttt{HELEN (Gauss)} consistently estimate the spectra of all materials that are close to the reference spectra in all the pixels under evaluation.

Fig. \ref{fig:real outlier} shows the detected outlier map and the estimated spectra of outlying pixels by \texttt{VOIMU}, \texttt{HELEN (Beta)} and \texttt{HELEN (Gauss)}. The identified pixels by these algorithms are located in similar regions. The spectra of these detected pixels are likely outliers, as they
appear at the boundary of two materials. It has been widely reported in the literature that nonlinear mixing effect happens around the shore of the Moffett data; see  \cite{lyu2021identifiability,wu2017stochastic,halimi2011nonlinear,fu2016robust}.

\medskip

\section{Conclusion}
In this work, we proposed a probabilistic HU-EV framework based on the MML principle. 
The proposed \texttt{HELEN} framework allows us to design efficient VI algorithms for HU-EV, with the flexibility to use a range of endmember priors.
Our design includes a patch-wise static endmember model that is the key to exploit the spatial smoothness and to avoid ill-posedness issues in using standard VI for HU-EV.
Using the proposed VI framework, we came up with detailed algorithm designs using the Beta and Gaussian priors, respectively.
In particular, the proposed \texttt{HELEN (Beta)} algorithm provides an efficient probabilistic inference solution for HU-EV under Beta priors of the endmembers, yet existing methods have to resort to computationally prohibitive sampling operations for the same purpose.
Our method also detects outliers that are ubiquitously observed in real-world HSIs.
Numerical results demonstrate the effectiveness of the proposed framework.

\bibliography{refs}

\end{document}

%% file: sym_marco.tex
\newcommand\bo{\ensuremath{{\bm o}}}

\newcommand\bx{\ensuremath{{\bm x}}}
\newcommand\by{\ensuremath{{\bm y}}}

\newcommand\bR{\ensuremath{{\bm R}}}

\newcommand\bX{\ensuremath{{\bm X}}}

\newcommand\bC{\ensuremath{{\bm C}}}

\newcommand\ba{\ensuremath{{\bm a}}}
\newcommand\bba{\ensuremath{\bar{\bm a}}}
\newcommand\bA{\ensuremath{{\bm A}}}
\newcommand\bbA{\ensuremath{\bar{\bm A}}}

\newcommand\bB{\ensuremath{{\bm B}}}

\newcommand\bmu{\ensuremath{{\bm \mu}}}
\newcommand\balp{\ensuremath{{\bm \alpha}}}

\newcommand\bD{\ensuremath{{\bm D}}}

\newcommand\bv{\ensuremath{{\bm v}}}
\newcommand\bSig{\ensuremath{{\bm \Sigma}}}

\newcommand\btheta{\ensuremath{{\bm \theta}}}

\newcommand\bphi{\ensuremath{{\bm \phi}}}

\newcommand\bY{\ensuremath{{\bm Y}}}
\newcommand\bV{\ensuremath{{\bm V}}}
\newcommand\bU{\ensuremath{{\bm U}}}
\newcommand\bs{\ensuremath{{\bm s}}}

\newcommand\bE{\ensuremath{{\bm E}}}

\newcommand\tbV{\ensuremath{\tilde{\bm V}}}

\newcommand\hba{\ensuremath{\widehat{{\bm a}}}}
\newcommand\hbA{\ensuremath{\widehat{{\bm A}}}}
\newcommand\hbs{\ensuremath{\widehat{{\bm s}}}}

\newcommand\bomg{\ensuremath{\bar{\omega}}}

\newcommand{\pout}{p_{\rm out}}
\newcommand{\pnom}{p_{\rm nom}}

\newcommand{\Rbb}{\mathbb{R}}

\newcommand{\setU}{\mathcal{U}}

\newcommand{\setN}{\mathcal{N}}
\newcommand{\setL}{\mathcal{L}}
\newcommand\bQ{\ensuremath{{\bm Q}}}

\newcommand\hl{\ensuremath{\widehat{ \ell}}}
\newcommand{\hL}{\widehat{\setL}}

\newcommand{\Exp}{\mathbb{E}}

\newcommand{\Diag}{\mathrm{Diag}}
\newcommand{\diag}{\mathrm{diag}}
\newcommand{\Beta}{\mathrm{Beta}}
\newcommand{\Dir}{\mathrm{Dir}}
\newcommand{\kl}{\mathrm{KL}}
\newcommand{\lognormal}{\mathrm{Lognormal}}

\newcommand{\bzero}{{\bm 0}}
\newcommand{\bone}{{\bm 1}}
\newcommand{\bI}{{\bm I}}

\newcommand{\setI}{\mathcal{I}}

\newcommand\indfn{{\mathbb 1}}

\newcommand\innerF[2]{\langle #1,#2 \rangle_F}

\newcommand\tr{\ensuremath{{\rm tr}}}

\errorcontextlines\maxdimen

\makeatletter
\newcommand*{\algrule}[1][\algorithmicindent]{\makebox[#1][l]{\hspace*{.5em}\thealgruleextra\vrule height \thealgruleheight depth \thealgruledepth}}%
\newcommand*{\thealgruleextra}{}
\newcommand*{\thealgruleheight}{.75\baselineskip}
\newcommand*{\thealgruledepth}{.25\baselineskip}

\newcount\ALG@printindent@tempcnta
\def\ALG@printindent{%
	\ifnum \theALG@nested>0
	\ifx\ALG@text\ALG@x@notext
	\else
	\unskip
	\addvspace{-1pt}
	\ALG@printindent@tempcnta=1
	\loop
	\algrule[\csname ALG@ind@\the\ALG@printindent@tempcnta\endcsname]%
	\advance \ALG@printindent@tempcnta 1
	\ifnum \ALG@printindent@tempcnta<\numexpr\theALG@nested+1\relax
	\repeat
	\fi
	\fi
}%
\usepackage{etoolbox}
\patchcmd{\ALG@doentity}{\noindent\hskip\ALG@tlm}{\ALG@printindent}{}{\errmessage{failed to patch}}
\makeatother

\newbox\statebox
\newcommand{\myState}[1]{%
	\setbox\statebox=\vbox{#1}%
	\edef\thealgruleheight{\dimexpr \the\ht\statebox+1pt\relax}%
	\edef\thealgruledepth{\dimexpr \the\dp\statebox+1pt\relax}%
	\ifdim\thealgruleheight<.75\baselineskip
	\def\thealgruleheight{\dimexpr .75\baselineskip+1pt\relax}%
	\fi
	\ifdim\thealgruledepth<.25\baselineskip
	\def\thealgruledepth{\dimexpr .25\baselineskip+1pt\relax}%
	\fi
	\State #1%
	\def\thealgruleheight{\dimexpr .75\baselineskip+1pt\relax}%
	\def\thealgruledepth{\dimexpr .25\baselineskip+1pt\relax}%
}